\pdfoutput=1

\documentclass[11pt]{article}

\usepackage[final]{acl}

\usepackage{times}
\usepackage{latexsym}

\usepackage[T1]{fontenc}

\usepackage[utf8]{inputenc}

\usepackage{microtype}

\usepackage{inconsolata}

\usepackage{graphicx}
\usepackage[utf8]{inputenc} 
\usepackage[T1]{fontenc}    
\usepackage{hyperref}       
\usepackage{url}            
\usepackage{booktabs}       
\usepackage{amsfonts}       
\usepackage{nicefrac}       
\usepackage{microtype}      
\usepackage{xcolor}         
\usepackage{colortbl}
\usepackage[breakable]{tcolorbox}
\usepackage{array}
\usepackage{color,soul}
\usepackage{siunitx}
\usepackage{multirow}
\usepackage[flushleft]{threeparttable}
\usepackage{subfigure}
\usepackage{caption}
\usepackage{subcaption}
\usepackage{geometry}
\usepackage{enumitem}
\usepackage{wrapfig}

\newcommand{\titlestyle}{\normalfont\large\bfseries}

\definecolor{darkgrey}{gray}{0.3}
\definecolor{lightgrey}{gray}{0.9}

\newtcolorbox{mybox}[2][]{%
    breakable,
    colback=lightgrey, 
    colframe=darkgrey, 
    fonttitle=\bfseries, 
    title=#2,
    #1
    }

\newtcolorbox{continuebox}{
    breakable,
    colback=lightgrey, 
    }

\newcommand{\data}{\emph{PISTOL}}

\title{How Data Inter-connectivity Shapes LLMs Unlearning:\\ A Structural Unlearning Perspective}

\author{
Xinchi Qiu\textsuperscript{1}\thanks{Equal contribution. Correspondence to Xinchi Qiu (\texttt{xq227@cam.ac.uk}) or William F. Shen (\texttt{fs604@cam.ac.uk}). Nicola Cancedda served in an advisor role.} \ William F. Shen\textsuperscript{1*}  Yihong Chen\textsuperscript{2}   \
Meghdad Kurmanji\textsuperscript{1} \\
\textbf{Nicola Cancedda\textsuperscript{3}  Pontus Stenetorp\textsuperscript{2}  Nicholas D.\ Lane\textsuperscript{1}} \\
\textsuperscript{1} Department of Computer Science and Technology, University of Cambridge\\
\textsuperscript{2} UCL Centre of Artificial Intelligence \ \textsuperscript{3} FAIR, Meta\\
}

\begin{document}
\maketitle

\begin{abstract}

While unlearning knowledge from large language models (LLMs) is receiving increasing attention, one important aspect remains unexplored. Existing approaches and benchmarks assume data points to-be-forgotten are independent, ignoring their inter-connectivity -- a fundamental characteristic of real-world data structures. In this paper, we propose \data{}, a method for compiling structural datasets. \data{} leverages the inherently structured nature of contractual relationships, offering several key benefits. First, it enables insights into the impact of structural data on unlearning effectiveness. Second, it provides precise and concise ground truths for clearer evaluation. Third, its attribute generation does not require input from pre-trained LLMs, mitigating confounding risks. Leveraging datasets synthesized using \data{}, we demonstrate how data inter-connectivity impacts LLM unlearning. Specifically \textbf{(a)} in both the pre-trained and fine-tuned models, unlearning difficulty increases as data inter-connectivity grows, \textbf{(b)} there is a positive correlation between the density of the knowledge graph and unlearning difficulty, and \textbf{(c)} when the to-be-forgotten data is skewed towards one domain, balancing retaining performance across all domains is challenging.


\end{abstract}

\section{Introduction}\label{sec:intro}

Large language models~(LLMs) have shown impressive capabilities in natural language generation.
However, their output is not always appropriate 
due to issues such as generating biased \citep{kotek2023gender, motoki2023more} or toxic content \citep{wen2023unveiling, bender2021dangers}, regurgitating personally identifiable information~(PII) \citep{nasr2023scalable, barrett2023identifying}, and \textit{hallucination} \citep{huang2023survey, xu2024hallucination}.

%

One straightforward way to mitigate these undesired behaviors is to \textit{retrain} the model on a new dataset which deletes `bad' data points that cause the unwanted behaviors. 
However, naively retraining 
is known to be highly inefficient ~\cite{lora,marchisio2023mini,zhang2023composing} due to significant computation cost and data requirements~\cite{chen2023improving}.
As an alternative, \textit{machine unlearning} (MU)~\cite{bourtoule2021machine,nguyen2022survey}, originally proposed for classification models, has been extended to remove the influence of undesirable data and model capabilities for LLMs
~\cite{zhang2023right, liu2024rethinking}.

\begin{figure*}[t] 
  \centering
  \includegraphics[width=0.7\textwidth]{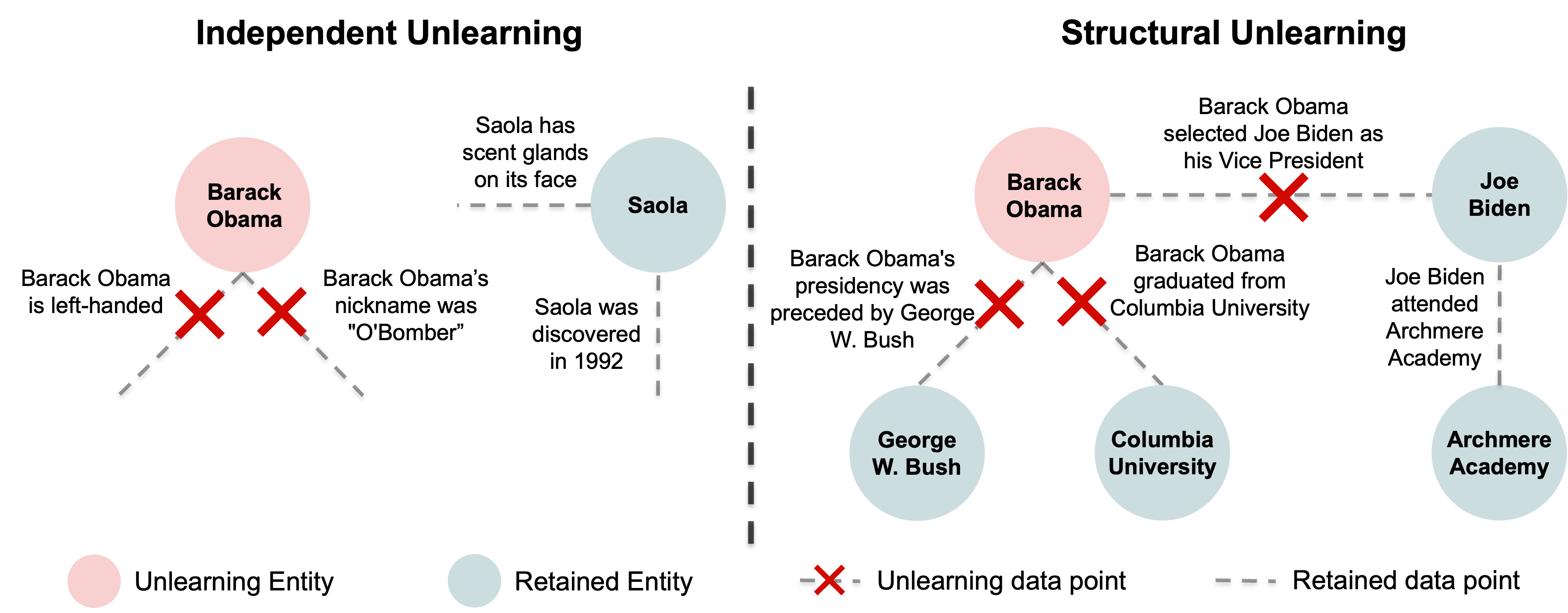}
  \captionsetup{font=small,labelfont=bf}
  \caption{Illustration of \textit{structural unlearning} (i.e., unlearning inter-connected data points within a structured dataset) versus \textit{independent unlearning} (i.e., unlearning isolated data points). As shown, when an entity revokes consent for its data to be used or exercises its \textit{`right to be forgotten'} (i.e., unlearning data points related to this entity), the degree of inter-connectivity between the unlearning entity and other entities will influence unlearning performance.}
  \label{fig:structural_unlearning_illu}
\end{figure*}

Despite being a promising direction, LLM unlearning remains nascent. 
%
Particularly, existing unlearning methods are often evaluated using datasets, such as TOFU \citep{maini2024tofu}, which primarily composed of independent entities. However, we observe that real data points (such as Wikipedia data) are rarely independent, they are often inter-connected, creating knowledge graphs with intricate topologies \citep{schneider2022decade}. As such, real-world unlearning usage extends beyond \textit{simple deletion} of independent data points from LLMs. Instead, it necessitates \textit{structural data deletion, which facilitates the comprehensive removal of `relational' data, irrespective of its inter-connectivity with other entities or its domain} (illustrated in Figure \ref{fig:structural_unlearning_illu}). Such graph-type inter-connected relationships among data points present a challenge in LLMs unlearning as forgetting one data point might impact the retaining/forgetting of others. Therefore, it is essential to assess the true effectiveness of existing LLMs unlearning algorithms in the presence of structural data points, and to develop realistic datasets that facilitate such research.


In this work, we are particularly interested in exploring two key research questions associated with structural LLM unlearning:

\textit{(1) How does data inter-connectivity impact the unlearning performance?} 
Some entities naturally appear more frequently in joint information with others. As such, an entity that revokes consent for its data to be used or exercises its \textit{`right to be forgotten'} may have varying levels of inter-connectivity with other entities in the dataset. An effective unlearning algorithm must robustly handle unlearning requests from entities which have varying levels of inter-connectivity while minimizing the need for manual intervention, such as extensive hyper-parameter tuning.

\textit{(2) How does unlearning data from a specific domain affect the retained model’s performance on data in the same versus different domain?} 
Another important aspect of structural unlearning is that unlearning requests may specifically target certain data domain rather than encompassing a mix of everything, as reflected by their proportional representation in the training set. 
For the first time, we investigate how such targeted unlearning affects outcomes, particularly examining whether it leads to uneven performance degradation on the retained data of both the same and different domains.

In summary: 
\begin{itemize}
    \vspace{-0.2cm}
    \item 
    We introduce \data{}, a novel dataset compilation pipeline, that reflects the structured nature of real-world knowledge and supports the study and evaluation of multi-scenario structural LLM unlearning.
    \vspace{-0.2cm}
    \item We demonstrate using \data{} datasets that the degree of inter-connectivity of a data point positively correlates with the difficulty of unlearning. We further show the same correlation also holds for pre-training data.
    \vspace{-0.2cm}
    \item We show unlearning data skewed towards a specific domain often leads to a more pronounced deterioration in the retained model’s performance on that same domain. 
    \vspace{-0.2cm}
    \item We compare two mainstream MU approaches, gradient ascent-based and preference optimization-based (PO)  methods, through qualitative and quantitative analysis. Our findings indicate that PO-based MU is more robust in handling structural unlearning.
    
\end{itemize}
\vspace{-0.3cm}
\section{Limitations of Existing Datasets}\label{sec:problem_of_tofu}

Existing LLM unlearning methods and datasets considered removing the influence of independent data points. TOFU \cite{maini2024tofu}, the recently created and commonly used dataset for LLM unlearning, is a dataset that consists of profiles of 200 fictitious authors, each has 20 question-answer pairs synthesized by GPT-4. Notably, each fictitious author is an isolated entity,  \textit{without apparent inter-connections} to other authors within the dataset. 

First, we investigate whether TOFU, or a slightly modified version of it, can be used to study structural unlearning. We observed that, except for two authors, weak entity relationships may be inferred within the original dataset in terms of authors' countries of birth. Among the 200 fictitious authors, 9 share the same country of birth, the U.S., whereas authors from countries like Zimbabwe and Portugal have no apparent connections with others.

Top half of Table \ref{tab:structured_tofu_new} shows the average Deviation Scores (DS) (see Sec. \ref{sec:evaluation_metrics}) 
of unlearning three randomly selected U.S. authors (who represent the highly inter-connected entity) compared to those from countries with a single author (who represent lowly inter-connected entity). 
The results reveal only a marginal difference in unlearning performance, suggesting that the model, when finetuned on the original TOFU dataset, has a limited appreciation of its weak inferred entity inter-connectivity.

Then, we further explore whether the original TOFU dataset can be modified to strengthen the inter-connectivity between certain authors. To this end, we modify the original dataset by introducing a more explicit knowledge type, based on personal relationships, than the country of birth. We call it as the new `Structured TOFU' dataset.
For each unlearning U.S. author, we select 5 other U.S. authors from the original dataset and create fictitious relationships (e.g., friends, coauthors, teachers) with the selected unlearning author. We then replace 10 out of the 20 QA pairs for the unlearning author with new QA pairs about his/her relationships with the other 5 authors. These new QA pairs were generated using GPT-4 and followed the same prompt format as described in the TOFU paper. We do not change QA pairs of other fictitious authors.

The lower half of Table \ref{tab:structured_tofu_new} presents the results of experiments conducted using the same procedure as for the original TOFU dataset. While introducing stronger inter-connectivity among authors slightly increases the unlearning impact compared to the original TOFU dataset, the overall magnitude of difference remains small. 


In conclusion, these findings, coupled with TOFU's inflexibility to assess the impact of data density and domain (see Sec. \ref{sec:contractdataset} and \ref{sec:eval_methods}), as well as other side effects (see Sec. \ref{sec:pipeline}), highlight the need for a novel dataset to better support the study and evaluation of LLM unlearning.

\begin{table}[t]
\centering
\captionsetup{font=small,labelfont=bf}
\caption{Deviation Scores ($\downarrow$) of unlearning a highly inter-connected entity (represented by US author) and a lowly inter-connected entity (represented by an author from a country with only one representative) for various unlearning methods using Llama2-7B.}
\label{tab:structured_tofu_new}
\scalebox{0.75}{
\begin{tabular}{lc |cc c}
\toprule
{} & {\textbf{Method}} & {\textbf{USA}} & \multicolumn{1}{c}{\shortstack{\textbf{Country} \\ \textbf{w/ 1 Author}}} & {\textbf{$\Delta$}} \\
\midrule
\multirow{5}{*}{\shortstack{\textbf{Original} \\ \textbf{TOFU}}} 
& GA  & 40.8 & 37.4 & 3.4 \\
& GD  & 40.7 & 36.0 & 4.7 \\
& UKL  & 73.3 & 71.0 & 2.2 \\
& DPO & 23.5 & 21.7 & 1.8 \\
& NPO & 45.4 & 42.3 & 3.1 \\

\midrule

\multirow{5}{*}{\shortstack{\textbf{Structured} \\ \textbf{TOFU}}} 
& GA  & 45.4 & 39.6 & 5.9 \\
& GD  & 44.8 & 39.3 & 5.5 \\
& UKL  & 70.3 & 68.4 & 1.9 \\
& DPO & 30.3 & 27.6 & 2.8 \\
& NPO & 45.5 & 42.1 & 3.3 \\

\bottomrule
\end{tabular}
}
\end{table}

\section{PISTOL Dataset}\label{sec:contractdataset}
\begin{figure*}[t!]
    \centering
    
    \includegraphics[clip,width=2.0\columnwidth]{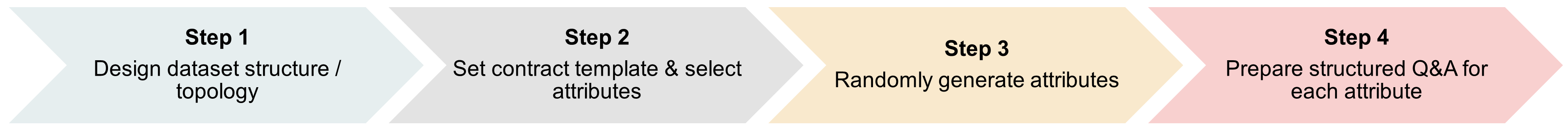}
     \captionsetup{font=small,labelfont=bf}
    \caption{Illustration of the Dataset Compilation Pipeline.}
    \label{fig:pipeline}
\end{figure*}

\begin{figure*}[hbt!]
    \setlength{\abovecaptionskip}{10pt} 
    \setlength{\belowcaptionskip}{-10pt} 
    \setlength{\floatsep}{5pt} 
    \setlength{\textfloatsep}{5pt} 
    \captionsetup{font=small,labelfont=bf}
    \centering
    \begin{minipage}[b]{0.4\textwidth}
        \centering
        \includegraphics[width=\textwidth]{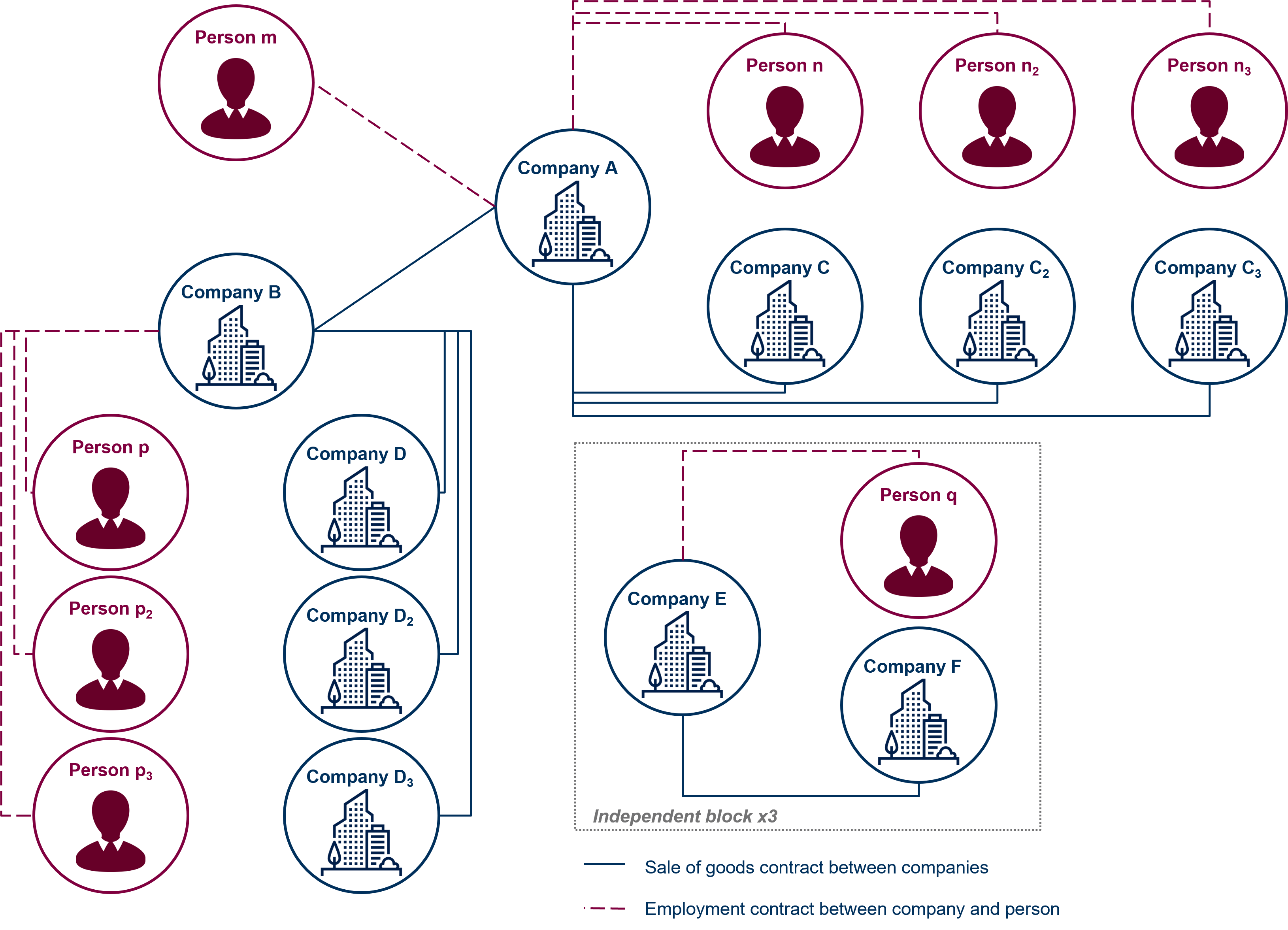}
        \caption*{\small (a)}
    \end{minipage}
    \begin{minipage}[b]{0.35\textwidth}
        \centering
        \includegraphics[width=\textwidth]{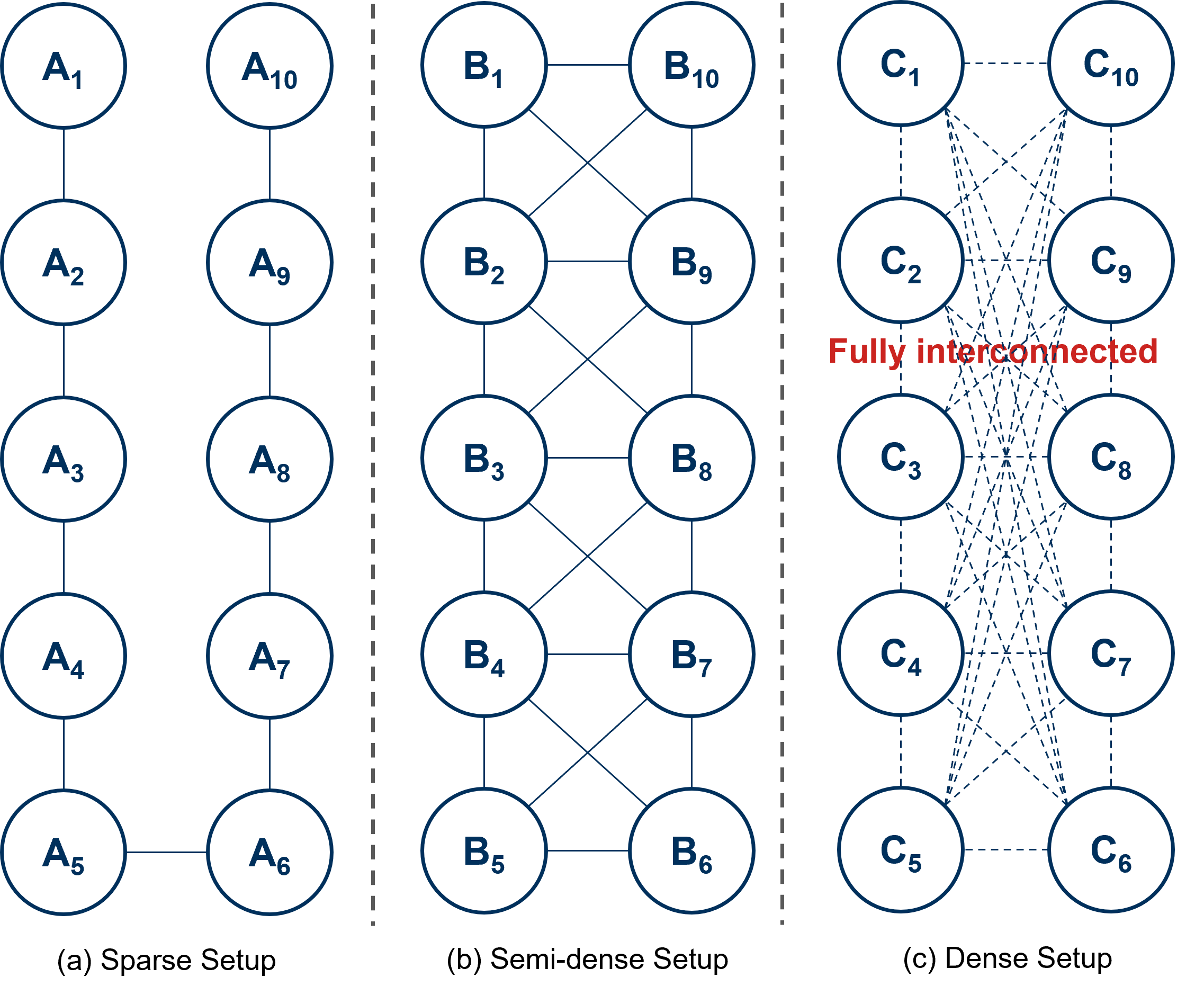}
        \caption*{\small (b)}
    \end{minipage}
    \caption{(a) and (b) illustrates the structure of Dataset 1 and 2 respectively.}
    \label{fig:dataset_structure}
\end{figure*}

In this section, we first introduce the novel dataset compilation \underline{pi}peline for \underline{st}ructural unlearning \underline{o}f \underline{L}LMs (\data{}) in Section \ref{sec:pipeline} and then two datasets generated by the \data{} in Section \ref{sec:dataset_pistol}.

\subsection{Dataset Compilation Pipeline} \label{sec:pipeline} 

To effectively reflect the structured nature of real-world data with well-defined and strong entity-level inter-connections, we specifically choose contracts as the basis of \data{} dataset. This choice brings additional benefits including (i) easy customization of the network structure to study the impact of specific target topologies, (ii) straightforward incorporation of side features for exploring other research topics (e.g., temporal features for studying unlearning of outdated data), (iii) a highly structured format compared to other knowledge types (e.g. news articles or books, etc.) for consistent measurement of unlearning performance. 
Despite their advantages, contract datasets are typically confidential, and high-quality public-domain sources remain scarce. Coupled with the necessity of distinguishing the dataset from the pre-training corpus for evaluation, we synthesize datasets based on real-world contract terms, ensuring both relevance and controlled evaluation of unlearning methods.

The pipeline for compiling datasets in a controlled manner is illustrated in Figure \ref{fig:pipeline}. 
Firstly, we craft the overall knowledge-graph structure, taking into account the structural variation of unlearning. 
Then we set the contract template, each with 20 attributes to be filled in. In our datasets, we focused on two ubiquitous types of contracts, sales of goods and employment contracts, owing to their more standardized structure in contrast to other highly customized agreements. 
Subsequently, we generate attributes in a \textbf{\textit{random}} manner, taking into account the dataset size. We randomly generate 6 letters and a suffix for a company name (e.g. Empblq LLC), 4 letters for the first name and the surname of a person (e.g. Jkeq Cyfz), 3 numbers, 6 letters and a street type for an address (e.g. 442 Rcvvyy Boulevard). Other attributes such as the signing date, contractual terms, and governing jurisdiction are also randomly generated. 
Finally, we prepare a QA pair for each attribute. QA pairs follow a consistent querying mechanism and have concise answers to allow systematic evaluations. Templates of both types of contracts and detailed QA of our sample datasets are provided in Appendix \ref{app:example}.

\data{} pipeline is distinctive from prior works such as TOFU from several perspectives. First, \data{} not only allows the synthesis of independent data points but also enables the design and creation of inter-connected data, reflecting this fundamental characteristic of real-world data structures.
Secondly, \data{} does \textit{not} depend on GPT or \textit{any} other pre-trained models for generating synthetic data, minimizing the risk of confounders. 
Thirdly, ground truth answers in \data{} are designed to be extremely precise and concise, containing only the targeted information for unlearning. This contrasts with TOFU which consists of open-ended QAs. As common evaluation metrics (e.g., ROUGE score, etc.) \citep{liu2024rethinking, romandini2024federated} compare generated responses with ground truth answers, the presence of tokens unrelated to unlearning may pollute the scores, resulting in inaccurate evaluation. By ensuring concise and precise QAs, \data{} mitigates the issue and provides a clearer measure of unlearning success.

\subsection{Dataset}\label{sec:dataset_pistol}
For the purpose of this research, we introduce two datasets compiled based on \data{} as below, with more details in Appendix \ref{app:sample_dataset}.



\vspace{-0.5em}
\paragraph{Dataset 1.} As depicted in Figure \ref{fig:dataset_structure}(a), has a structure of $G(24, 20)$ topology (a graph with $24$ nodes and $20$ edges).
%
%
It contains two data domains -- sales contracts between companies or employment contracts between companies and individuals. 
Dataset 1 has a symmetric structure, allowing for the controlled isolation of topological impact when evaluating multiple structural features, such as entity inter-connectivity and data domain. 

\vspace{-0.5em}
\paragraph{Dataset 2.} 
Despite its advantages, the symmetric structure of Dataset 1 limits topological variations when sampling different unlearning edges, providing only a `local view' of the impact of data inter-connectivity. To achieve greater sampling variability and verify our findings with respect to data inter-connectivity under the Dataset 1, we go beyond its symmetric structure and introduce a Dataset 2 as depicted in Figure \ref{fig:dataset_structure}(b). 

Dataset 2 comprises 3 sub-graphs -- each has 10 nodes but different inter-connectivity (edges). The most data \textit{sparse} sub-graph has 9 edges (i.e. connecting each node by a \textit{chain}). On the opposite end, the most data \textit{dense} sub-graph has 45 edges (i.e. nodes are \textit{fully connected}). The sub-graph in between is \textit{semi-dense} and has 21 edges. As Dataset 2 is designed for evaluating data inter-connectivity from a `global view', we isolate the data domain and only have sales contracts between companies.

\section{Evaluation Setup}
\label{sec:evaluation_metrics}
In this section, we introduce the experimental and evaluation setup and evaluation methods for the new structural LLMs unlearning considerations.

\vspace{-0.2cm}
\paragraph{Metrics.} We measure \emph{unlearning effectiveness} through two key aspects: forget efficacy, which captures how much the model’s outputs diverge from the forget set, and model utility, which reflects the preserved performance on data outside the forget set. Since these objectives are equally critical, we use \textit{Deviation Score (DS)} $= 100 \times \sqrt{ \text{ROUGE1}_\text{forget}^2 + (1 -\text{ROUGE1}_\text{retain})^2}$ to measure the Euclidean distances of forget efficacy and model utility to their respective ideal state. A lower DS indicates more effective unlearning, signifying a closer approach to the optimal state -- where the model outputs no information from the forget set while maintaining full accuracy on the retained data. In contrast, a higher DS reflects poorer unlearning, suggesting a weaker distinction between forget and retained knowledge. More details and other supplementary metrics, including the original ROUGE1 scores, MRR and the Top Hit Rate, can be found in Appendix \ref{app:evalmetricsdetails}.

\vspace{-0.2cm}
\paragraph{Unlearning baselines}
We experiment with three gradient-based methods: Gradient Ascent (GA) \citep{jang2022knowledge, yao2023large}, Gradient Difference (GD) \citep{liu2022continual} and GA with KL-divergence regularization (UKL), as well as two preference optimization(PO)-based methods: Direct Preference Optimization (DPO) \citep{rafailov2024direct} and Negative Preference Optimization (NPO) \citep{zhang2024negative}. Given the nascence of the field, existing unlearning methods often lack robustness. However, these methods represent the current mainstream and serve well to demonstrate the impact of structural datasets while inspiring further research. 

\vspace{-0.2cm}
\paragraph{Base models.} We evaluate all baseline methods using the current widely adopted language models Llama2-7B \cite{llama}, Gemma-7B \cite{gemma} and Mistral-7B \cite{jiang2023mistral}. We evaluated learning rates between $1\times 10^{-6}$ and $5\times 10^{-5}$ during unlearning and found that all methods are highly sensitive to learning rate and batch size selection. Since successful unlearning must preserve model utility, we enforce a performance threshold for ROUGE1 of the retained dataset and select the learning rate that maximizes forgetting. We conduct extensive learning rate tuning for each method and model, as all baseline unlearning methods demonstrate significant sensitivity to it -- a characteristic that poses challenges for their practical implementation. To ensure a fair comparison, we hold the learning rate constant when testing unlearning performance across different data (i.e., data with varying degrees of inter-connectivity or from different domains), provided the method and base model remained unchanged. Details of fine-tuning and unlearning methods are included in the Appendix \ref{sec:finetuneing} and \ref{sec:unlearningmethod}.

\subsection{Evaluation Methods}\label{sec:eval_methods}
\paragraph{Impact of data inter-connectivity.}
We define the inter-connectivity of a data point (edge) as the total degree of the vertices (entities) it connects: $deg(e_i) = \sum_{v\in e_i} deg(v)-1$. A higher degree indicates greater inter-connectivity. In Dataset 1, companies \( A \) and \( B \) have signed 8 and 7 contracts, respectively (including the contract between \( A \) and \( B \)). Thus, edge \( AB \), with a degree of 14, exhibits higher inter-connectivity than edge \( AC \), with a degree of 8. To assess the impact of inter-connectivity, we compare the outcomes of unlearning contracts in the same domain but with different inter-connectivity levels (e.g., unlearning sales contracts \( AB \) and \( AC \) respectively).

To assess from a `global' data inter-connectivity perspective (i.e., \textit{the impact of data density on unlearning performance}), we exploit the asymmetric nature of Dataset 2 by randomly selecting an edge to forget in each sub-graph with different data densities. We perform experiments three times and take average of the results. 

\paragraph{Impact of data domain.}
We contrast unlearning a sales contract between company $A$ and company $C$ with unlearning an employment contract between company $A$ and individual $n$. Model utility post-unlearning is evaluated on the independent retained sales edge ($EF$) and employment edge ($Eq$), respectively. As both $EF$ and $Eq$ belong to an isolated sub-graph, this design isolates the effect of data domain from confounding factors such as data inter-connectivity.

\section{Results}\label{sec:results}
In this section, we report the experimental results and discuss our key findings.

\begin{tcolorbox}[
    colback=gray!10,  
    colframe=gray!10, 
    boxrule=0pt,      
    sharp corners,    
    width=\columnwidth 
]
\textbf{\textit{Takeaway 1:}} The greater the degree of inter-connectivity a data
point has, the more challenging it becomes to unlearn the data.
\end{tcolorbox}
\begin{table*}[ht]
\centering
\captionsetup{font=small,labelfont=bf}
\caption{Performance of unlearning a highly inter-connected data ($AB$) versus a lowly inter-connected data ($AC$) in Dataset 1 across various unlearning baselines and models. Deviation Scores ($\downarrow$) of unlearning $AB$ is consistently higher than those of $AC$, demonstrating a positive correlation between the level of data inter-connectivity and unlearning difficulty.}
\label{tab:ab_ac}
\scalebox{0.63}{
\begin{tabular}{ll |cc c|cc c|cc c}
\toprule
\multicolumn{2}{c|}{} &
\multicolumn{3}{c|}{\textbf{Llama2}} & 
\multicolumn{3}{c|}{\textbf{Mistral}} & \multicolumn{3}{c}{\textbf{Gemma}} \\
\midrule
{\textbf{Forget}} &
{\textbf{Forget}} & 
{\textbf{Forget}} &
{\textbf{Retain}} & 
{\textbf{Deviation}} & 
{\textbf{Forget}} &
{\textbf{Retain}} & 
{\textbf{Deviation}} & 
{\textbf{Forget}} &
{\textbf{Retain}} & 
{\textbf{Deviation}} \\

{\textbf{Data}} &
{\textbf{Method}} & 
{\textbf{ROUGE1 ($\downarrow$)}} & 
{\textbf{ROUGE1 ($\uparrow$)}} &
\textbf{Score ($\downarrow$)} & 
{\textbf{ROUGE1 ($\downarrow$)}} & 
{\textbf{ROUGE1 ($\uparrow$)}} &
\textbf{Score ($\downarrow$)} & 
{\textbf{ROUGE1 ($\downarrow$)}} & 
{\textbf{ROUGE1 ($\uparrow$)}} &
\textbf{Score ($\downarrow$)} \\

\midrule
\multirow{5}{*}{\textbf{AB}} 
& GA  & 0.521 $\pm$ 0.050 & 0.845 $\pm$ 0.043 & 54.4 & 0.325 $\pm$ 0.029 & 0.851 $\pm$ 0.010 & 35.8 & 0.563 $\pm$ 0.024 & 0.879 $\pm$ 0.006 & 57.6 \\
& GD  & 0.654 $\pm$ 0.029 & 0.944 $\pm$ 0.003 & 65.6 & 0.347 $\pm$ 0.075 & 0.906 $\pm$ 0.021 & 36.0 & 0.319 $\pm$ 0.080 & 0.844 $\pm$ 0.046 & 35.5 \\
& UKL  & 0.700 $\pm$ 0.050 & 0.936 $\pm$ 0.036 & 70.3 & 0.667 $\pm$ 0.052 & 0.969 $\pm$ 0.014 & 66.8 & 0.730 $\pm$ 0.062 & 0.916 $\pm$ 0.006 & 73.5 \\
& DPO & 0.300 $\pm$ 0.000 & 0.902 $\pm$ 0.023 & 31.6 & 0.150 $\pm$ 0.050 & 0.878 $\pm$ 0.019 & 19.3 & 0.193 $\pm$ 0.001 & 0.785 $\pm$ 0.033 & 28.9 \\
& NPO & 0.380 $\pm$ 0.000 & 0.880 $\pm$ 0.006 & 39.8 & 0.312 $\pm$ 0.025 & 0.818 $\pm$ 0.001 & 36.1 & 0.328 $\pm$ 0.025 & 0.845 $\pm$ 0.028 & 39.4 \\
\midrule
\multirow{5}{*}{\textbf{AC}} 
& GA  & 0.267 $\pm$ 0.029 & 0.805 $\pm$ 0.005 & 33.1 & 0.184 $\pm$ 0.058 & 0.890 $\pm$ 0.010 & 21.4 & 0.502 $\pm$ 0.021 & 0.917 $\pm$ 0.005 & 50.9 \\
& GD  & 0.283 $\pm$ 0.029 & 0.920 $\pm$ 0.005 & 29.4 & 0.261 $\pm$ 0.077 & 0.959 $\pm$ 0.010 & 26.4 & 0.309 $\pm$ 0.008 & 0.893 $\pm$ 0.006 & 32.7 \\
& UKL  & 0.505 $\pm$ 0.256 & 0.793 $\pm$ 0.165 & 54.6 & 0.739 $\pm$ 0.010 & 0.968 $\pm$ 0.008 & 74.0 & 0.600 $\pm$ 0.068 & 0.943 $\pm$ 0.003 & 60.3 \\
& DPO & 0.242 $\pm$ 0.000 & 0.933 $\pm$ 0.008 & 25.1 & 0.103 $\pm$ 0.000 & 0.949 $\pm$ 0.004 & 11.5 & 0.180 $\pm$ 0.050 & 0.808 $\pm$ 0.025 & 26.3 \\
& NPO & 0.363 $\pm$ 0.035 & 0.897 $\pm$ 0.002 & 37.8 & 0.300 $\pm$ 0.010 & 0.902 $\pm$ 0.006 & 31.5 & 0.374 $\pm$ 0.016 & 0.916 $\pm$ 0.006 & 38.4 \\
\bottomrule
\end{tabular}
}
\end{table*}

Table \ref{tab:ab_ac} compares the results of unlearning a highly inter-connected data point ($AB$) versus a less inter-connected one ($AC$) using various baseline methods and models. The results show that the Deviation Scores (DS) for unlearning the higher inter-connected data $AB$ are consistently higher across all unlearning methods, with the GD method on Llama2-7B exhibiting the largest gap -- up to $2.2$x higher than the score for unlearning the less inter-connected data $AC$\footnote{A higher separation of DS scores, compared to evaluations using LlamA2-7B model on the Structured TOFU dataset in Sec. \ref{sec:problem_of_tofu}, further highlights the clearer inter-connectivity in \data{} -- one of its key advantages.}. Higher DS for unlearning higher inter-connected data indicate weaker separation between output accuracy on the forget and retain sets, and thus a more challenging unlearning process. Additional results for other metrics can be found in Appendix \ref{app:additionalres}.

The results also show that the difference is particularly pronounced when unlearning with the GA and GD methods, whereas data inter-connectivity has a smaller impact on the UKL method. However, the latter is primarily due to the overall poor performance of UKL. UKL is highly sensitive to learning rate adjustments, and to maintain a retain ROUGE1 score at a sensible level for preserving retained model utility, the forget ROUGE1 remains above $0.5$ -- an unacceptably high value, indicating a failure to effectively unlearn. This phenomenon is likely attributable to the design of UKL, which aims to reduce distribution shifts before and after the forgetting process. By preserving relational information between data points, UKL not only makes it more challenging to forget less inter-connected information ($AC$ edge) but also increases the difficulty of forgetting other data points.

Preference optimization (PO)-based methods show less sensitivity to data inter-connectivity despite being highly sensitive to learning rate changes. One reason for this could be that while the GA-based methods involve gradient ascent on the forget samples, PO-based methods continues to perform gradient descent, following the gradient of the forget set paired with negative examples such as `I don't know'. This suggests that PO-based methods may be more robust for structural unlearning.

%

\begin{tcolorbox}[
    colback=gray!10,  
    colframe=gray!10, 
    boxrule=0pt,      
    sharp corners,    
    width=\columnwidth 
]
\textbf{\textit{Takeaway 2:}} The greater density a knowledge
graph has, the more challenging it becomes to unlearn the data within the graph.
\end{tcolorbox}
\begin{table*}[ht]
\centering
\captionsetup{font=small,labelfont=bf}
\caption{Performance of unlearning data from sub-graphs of various densities in Dataset 2 across various unlearning baselines and models. Deviation Scores ($\downarrow$) of unlearning data from a dense sub-graph is consistently higher than those from less dense sub-graphs, demonstrating a positive correlation between density of knowledge graph and and unlearning difficulty.}
\label{tab:dataset2_res}
\scalebox{0.61}{
\begin{tabular}{ll |ccc |ccc |ccc}
\toprule
\multicolumn{2}{c|}{} &
\multicolumn{3}{c|}{\textbf{Llama2}} & 
\multicolumn{3}{c|}{\textbf{Mistral}} & \multicolumn{3}{c}{\textbf{Gemma}} \\
\midrule
{\textbf{Forget}} &
{\textbf{Forget}} & 
{\textbf{Forget}} &
{\textbf{Retain}} & 
{\textbf{Deviation}} & 
{\textbf{Forget}} &
{\textbf{Retain}} & 
{\textbf{Deviation}} & 
{\textbf{Forget}} &
{\textbf{Retain}} & 
{\textbf{Deviation}} \\

{\textbf{Data}} &
{\textbf{Method}} & 
{\textbf{ROUGE1 ($\downarrow$)}} & 
{\textbf{ROUGE1 ($\uparrow$)}} &
\textbf{Score ($\downarrow$)} & 
{\textbf{ROUGE1 ($\downarrow$)}} & 
{\textbf{ROUGE1 ($\uparrow$)}} &
\textbf{Score ($\downarrow$)} & 
{\textbf{ROUGE1 ($\downarrow$)}} & 
{\textbf{ROUGE1 ($\uparrow$)}} &
\textbf{Score ($\downarrow$)} \\
\midrule
\multirow{5}{*}{\textbf{Dense}} 
& GA  & 0.576 $\pm$ 0.093 & 0.997 $\pm$ 0.003 & 57.6 & 0.403 $\pm$ 0.096 & 0.961 $\pm$ 0.035 & 40.5 & 0.412 $\pm$ 0.025 & 0.944 $\pm$ 0.021 & 41.6 \\
& GD  & 0.623 $\pm$ 0.115 & 0.999 $\pm$ 0.002 & 62.3 & 0.435 $\pm$ 0.079 & 0.968 $\pm$ 0.034 & 43.6 & 0.369 $\pm$ 0.056 & 0.956 $\pm$ 0.027 & 37.2 \\
& UKL  & 0.302 $\pm$ 0.080 & 0.528 $\pm$ 0.110 & 56.0 & 0.613 $\pm$ 0.079 & 0.971 $\pm$ 0.004 & 61.3  & 0.628 $\pm$ 0.126 & 0.872 $\pm$ 0.054 & 64.1 \\
& DPO & 0.230 $\pm$ 0.057 & 0.998 $\pm$ 0.004 & 23.0 & 0.050 $\pm$ 0.061 & 0.993 $\pm$ 0.006 & 5.0 & 0.087 $\pm$ 0.013 & 0.855 $\pm$ 0.005 & 16.9 \\
& NPO & 0.099 $\pm$ 0.039 & 0.818 $\pm$ 0.037 & 20.8 & 0.408 $\pm$ 0.017 & 0.988 $\pm$ 0.013 & 40.9 & 0.386 $\pm$ 0.046 & 0.967 $\pm$ 0.013 & 38.8 \\
\midrule
\multirow{5}{*}{\textbf{Semi-Dense}} 
& GA  & 0.469 $\pm$ 0.055 & 0.999 $\pm$ 0.002 & 46.9 & 0.380 $\pm$ 0.087 & 0.977 $\pm$ 0.012 & 38.1 & 0.360 $\pm$ 0.073 & 0.940 $\pm$ 0.003 & 36.5 \\
& GD  & 0.598 $\pm$ 0.074 & 1.000 $\pm$ 0.000 & 59.8 & 0.377 $\pm$ 0.089 & 0.983 $\pm$ 0.007 & 37.7 & 0.277 $\pm$ 0.073 & 0.901 $\pm$ 0.008 & 29.4 \\
& UKL  & 0.258 $\pm$ 0.086 & 0.405 $\pm$ 0.141 & 64.9 & 0.450 $\pm$ 0.100 & 0.886 $\pm$ 0.050 & 46.4 & 0.775 $\pm$ 0.075 & 0.960 $\pm$ 0.020 & 77.6 \\
& DPO & 0.220 $\pm$ 0.055 & 0.996 $\pm$ 0.004 & 22.0 & 0.075 $\pm$ 0.056 & 0.991 $\pm$ 0.004 & 7.6 & 0.031 $\pm$ 0.019 & 0.846 $\pm$ 0.021 & 15.7 \\
& NPO & 0.177 $\pm$ 0.064 & 0.909  $\pm$ 0.016 & 21.2 & 0.399 $\pm$ 0.077 & 0.977 $\pm$ 0.009 & 40.0 & 0.350 $\pm$ 0.037 & 0.938 $\pm$ 0.010 & 35.6 \\
\midrule
\multirow{5}{*}{\textbf{Sparse}} 
& GA  & 0.430 $\pm$ 0.104 & 0.994 $\pm$ 0.004 & 43.0 & 0.268 $\pm$ 0.098 & 0.938 $\pm$ 0.027 & 27.5 & 0.290 $\pm$ 0.077 & 0.893 $\pm$ 0.022 & 30.9 \\
& GD  & 0.538 $\pm$ 0.144 & 0.997 $\pm$ 0.003 & 53.8 & 0.272 $\pm$ 0.102 & 0.957 $\pm$ 0.017 & 27.5 & 0.234 $\pm$ 0.020 & 0.826 $\pm$ 0.046 & 29.1 \\
& UKL  & 0.197 $\pm$ 0.085 & 0.478 $\pm$ 0.092 & 55.8 & 0.407 $\pm$ 0.043 & 0.887 $\pm$ 0.029 & 42.5 & 0.550 $\pm$ 0.450 & 0.827 $\pm$ 0.223 & 60.5 \\
& DPO & 0.220 $\pm$ 0.091 & 0.980 $\pm$ 0.023 & 22.1 & 0.030 $\pm$ 0.027 & 0.976 $\pm$ 0.010 & 3.8 & 0.031 $\pm$ 0.019 & 0.831 $\pm$ 0.004 & 17.1 \\
& NPO & 0.099 $\pm$ 0.039 & 0.818 $\pm$ 0.037 & 20.8 & 0.231 $\pm$ 0.010 & 0.957 $\pm$ 0.005 & 23.5 & 0.236 $\pm$ 0.022 & 0.871 $\pm$ 0.026 & 26.9 \\

\bottomrule
\end{tabular}
}
\end{table*}

Table \ref{tab:dataset2_res} compares the results of unlearning data from sub-graphs of different densities in Dataset 2 across various baselines and methods. The results indicate that unlearning becomes more challenging as knowledge graph becomes denser. The DS for the GA method using Llama2-7B increases from $43.0$ to $57.6$ as the average data inter-connectivity (i.e., knowledge density) increases from a sparse to a dense sub-graph. Similar upward trends are observed for the GD method. As before, the UKL method proves to be the least effective, consistently yielding poor unlearning performance in all settings. Consistent with the results for Dataset 1, the DPO method appears the most robust to variations in knowledge density, as evidenced by its relatively stable DS across sub-graphs with different densities in Dataset 2, with differences remaining within the statistical margin of error.

As explained in Section \ref{sec:dataset_pistol}, randomly sampling unlearning data in Dataset 2 allows us to evaluate the impact of structured dataset beyond the perspective of an individual entity’s degree of connectivity. Instead, it enables assessment from a `global view' of knowledge graph density. These findings not only support our earlier observations in \textit{Takeaway 1}, but also confirm a positive correlation between knowledge graph density and unlearning difficulty.

\begin{tcolorbox}[
    colback=gray!10,  
    colframe=gray!10, 
    boxrule=0pt,      
    sharp corners,    
    width=\columnwidth 
]
\textbf{\textit{Takeaway 3:}} Unlearning a specific type of data may lead to greater performance deterioration on data of the same type compared to data of a different type.
\end{tcolorbox}
\begin{table*}[ht]
\centering
\captionsetup{font=small,labelfont=bf}
\caption{Performance of unlearning data of different domains in Dataset 1 across various unlearning baselines and models. When the targeted unlearning data is a sales contract ($AC$), the ROUGE1 score of retained data within the same domain (i.e., sales contract data $EF$) drops more than the ROUGE1 score of retained data from a different domain (i.e., employment contract data $Eq$). This illustrates that unlearning data skewed toward a specific domain would lead to a more pronounced deterioration in the retained model’s performance on that same domain.}
\label{tab:takeaway3}
\scalebox{0.55}{
\begin{tabular}{ll |ccc |ccc |ccc}
\toprule
\multicolumn{2}{c|}{} &
\multicolumn{3}{c|}{\textbf{Llama2}} & 
\multicolumn{3}{c|}{\textbf{Mistral}} & \multicolumn{3}{c}{\textbf{Gemma}} \\
\midrule
&
& 
&
{\textbf{Ind. Retained}} & 
{\textbf{Ind. Retained}} & 
&
{\textbf{Ind. Retained}} & 
{\textbf{Ind. Retained}} & 
&
{\textbf{Ind. Retained}} & 
{\textbf{Ind. Retained}} \\

{\textbf{Forget}} &
{\textbf{Forget}} & 
{\textbf{Forget}} &
{\textbf{Sales Set ($EF$)}} & 
{\textbf{Emp. Set ($Eq$)}} & 
{\textbf{Forget}} &
{\textbf{Sales Set ($EF$)}} & 
{\textbf{Emp. Set ($Eq$)}} & 
{\textbf{Forget}} &
{\textbf{Sales Set ($EF$)}} & 
{\textbf{Emp. Set ($Eq$)}} \\

{\textbf{Data}} &
{\textbf{Method}} & 
{\textbf{ROUGE1 ($\downarrow$)}} & 
{\textbf{ROUGE1 ($\uparrow$)}} &
\textbf{Score ($\uparrow$)} & 
{\textbf{ROUGE1 ($\downarrow$)}} & 
{\textbf{ROUGE1 ($\uparrow$)}} &
\textbf{Score ($\uparrow$)} & 
{\textbf{ROUGE1 ($\downarrow$)}} & 
{\textbf{ROUGE1 ($\uparrow$)}} &
\textbf{Score ($\uparrow$)} \\

\midrule
\multirow{5}{*}{\textbf{AC (Sales)}} 
& GA  & 0.267 $\pm$ 0.029 & 0.772 $\pm$ 0.009 & 0.955 $\pm$ 0.010 & 0.184 $\pm$ 0.058 & 0.905 $\pm$ 0.010 & 0.983 $\pm$ 0.000 & 0.502 $\pm$ 0.021 & 0.946 $\pm$ 0.004 & 0.994 $\pm$ 0.004 \\
& GD  & 0.283 $\pm$ 0.029 & 0.911 $\pm$ 0.010 & 0.961 $\pm$ 0.000 & 0.261 $\pm$ 0.077 & 0.963 $\pm$ 0.003 & 1.000 $\pm$ 0.000 & 0.309 $\pm$ 0.008 & 0.888 $\pm$ 0.020 & 0.990 $\pm$ 0.004 \\
& KL  & 0.505 $\pm$ 0.256 & 0.772 $\pm$ 0.186 & 0.821 $\pm$ 0.119 & 0.739 $\pm$ 0.010 & 1.000 $\pm$ 0.000 & 1.000 $\pm$ 0.000 & 0.600 $\pm$ 0.068 & 0.939 $\pm$ 0.008 & 0.991 $\pm$ 0.007 \\
& DPO & 0.242 $\pm$ 0.000 & 0.939 $\pm$ 0.010 & 0.983 $\pm$ 0.000 & 0.103 $\pm$ 0.000 & 0.967 $\pm$ 0.017 & 0.983 $\pm$ 0.000 & 0.180 $\pm$ 0.050 & 0.767 $\pm$ 0.050 & 0.967 $\pm$ 0.017 \\
& NPO & 0.363 $\pm$ 0.035 & 0.928 $\pm$ 0.008 & 0.967 $\pm$ 0.000 & 0.300 $\pm$ 0.010 & 0.881 $\pm$ 0.008 & 0.983 $\pm$ 0.000 & 0.374 $\pm$ 0.016 & 0.961 $\pm$ 0.008 & 0.998 $\pm$ 0.003 \\
\midrule
\multirow{5}{*}{\textbf{An (Emp.)}} 
& GA  & 0.263 $\pm$ 0.000 & 0.950 $\pm$ 0.029 & 0.744 $\pm$ 0.013 & 0.345 $\pm$ 0.110 & 0.904 $\pm$ 0.013 & 0.839 $\pm$ 0.029 & 0.300 $\pm$ 0.025 & 0.994 $\pm$ 0.006 & 0.742 $\pm$ 0.000 \\
& GD  & 0.296 $\pm$ 0.029 & 0.983 $\pm$ 0.017 & 0.886 $\pm$ 0.013 & 0.286 $\pm$ 0.035 & 0.933 $\pm$ 0.000 & 0.950 $\pm$ 0.010 & 0.283 $\pm$ 0.05 & 0.978 $\pm$ 0.006 & 0.835 $\pm$ 0.026 \\
& KL  & 0.241 $\pm$ 0.029 & 0.819 $\pm$ 0.003 & 0.844 $\pm$ 0.067 & 0.828 $\pm$ 0.009 & 1.000 $\pm$ 0.000 & 0.978 $\pm$ 0.009 & 0.273 $\pm$ 0.015 & 0.936 $\pm$ 0.031 & 0.737 $\pm$ 0.001 \\
& DPO & 0.075 $\pm$ 0.025 & 0.989 $\pm$ 0.010 & 0.939 $\pm$ 0.010 & 0.010 $\pm$ 0.000 & 0.983 $\pm$ 0.000 & 0.978 $\pm$ 0.009 & 0.037 $\pm$ 0.009 & 0.928 $\pm$ 0.016 & 0.856 $\pm$ 0.008 \\
& NPO & 0.347 $\pm$ 0.016 & 0.972 $\pm$ 0.008 & 1.000 $\pm$ 0.000 & 0.037 $\pm$ 0.037 & 0.842 $\pm$ 0.025 & 0.631 $\pm$ 0.004 & 0.300 $\pm$ 0.025 & 0.986 $\pm$ 0.003 & 0.742 $\pm$ 0.000 \\
\bottomrule
\end{tabular}
}
\end{table*}

The structured nature of real-world data implies that information can belong to broader categorical domains. We seek to address how an unlearned model, after applying existing unlearning methods, performs on retained data within the same domain as the unlearned data compared to data from different domains. 

As shown in Table \ref{tab:takeaway3}, the results demonstrate that unlearning data skewed toward a specific domain often leads to a more pronounced deterioration in the model’s performance on retained data within that same domain. Specifically, ROUGE1 scores for independent retained sales contracts are lower than those for independent retained employment contracts when the unlearning data $AC$ is also a sales contract. Conversely, when the unlearning data is switched to $An$, an employment contract, the ROUGE1 scores for independent retained employment contracts decrease more than those for independent retained sales contracts. This phenomenon is particularly pronounced when using the GA method, highlighting its greater lack of robustness in handling unlearning data skewed toward a specific domain compared to other methods. 

\section{Inter-connectivity of Pre-training Data}\label{sec:interconnectivity_results}

\begin{table*}[ht]
\centering
\captionsetup{font=small,labelfont=bf}
\caption{Performance of unlearning highly inter-connected pre-training data (`Goldman Sachs') and less inter-connected data (`William Shakespeare') for various unlearning baselines and models. Higher Deviation Scores ($\downarrow$) observed for `Goldman Sachs' indicate that the challenge of unlearning highly inter-connected data extends to pre-training data, supporting our finding that more inter-connected data is harder to forget, consistent with results for synthetic data.}
\label{tab:GSWS_rouge1}
\scalebox{0.6}{
\begin{tabular}{ll |ccc |ccc |ccc}
\toprule
\multicolumn{2}{c|}{} &
\multicolumn{3}{c|}{\textbf{Llama2}} & 
\multicolumn{3}{c|}{\textbf{Mistral}} & \multicolumn{3}{c}{\textbf{Gemma}} \\
\midrule
{\textbf{Forget}} &
{\textbf{Forget}} & 
{\textbf{Forget}} &
{\textbf{Retain}} & 
{\textbf{Deviation}} & 
{\textbf{Forget}} &
{\textbf{Retain}} & 
{\textbf{Deviation}} & 
{\textbf{Forget}} &
{\textbf{Retain}} & 
{\textbf{Deviation}} \\

{\textbf{Data}} &
{\textbf{Method}} & 
{\textbf{ROUGE1 ($\downarrow$)}} & 
{\textbf{ROUGE1 ($\uparrow$)}} &
\textbf{Score ($\downarrow$)} & 
{\textbf{ROUGE1 ($\downarrow$)}} & 
{\textbf{ROUGE1 ($\uparrow$)}} &
\textbf{Score ($\downarrow$)} & 
{\textbf{ROUGE1 ($\downarrow$)}} & 
{\textbf{ROUGE1 ($\uparrow$)}} &
\textbf{Score ($\downarrow$)} \\
\midrule
\multirow{5}{*}{\shortstack[l]{\textbf{Goldman} \\ \textbf{Sachs}}} 
& GA  & 0.621 $\pm$ 0.106 & 1.000 $\pm$ 0.000 & 62.1 & 0.451 $\pm$ 0.120 & 0.995 $\pm$ 0.004 & 45.1 & 0.563 $\pm$ 0.039	& 0.975 $\pm$ 0.019 & 56.3 \\
& GD  & 0.704 $\pm$ 0.053 & 1.000 $\pm$ 0.000 & 70.4 & 0.485 $\pm$ 0.097 & 0.997 $\pm$ 0.002 & 48.5 & 0.509 $\pm$ 0.118 & 0.982 $\pm$ 0.019 & 50.9\\
& KL  & 0.357 $\pm$ 0.101 & 0.660 $\pm$ 0.081 & 49.3 & 0.732 $\pm$ 0.163 & 1.000 $\pm$ 0.000 & 73.2 & 0.722 $\pm$ 0.238 & 0.972 $\pm$ 0.026 & 72.3 \\
& DPO & 0.387 $\pm$ 0.193 & 1.000 $\pm$ 0.000 & 38.7 & 0.037 $\pm$ 0.027 & 1.000 $\pm$ 0.000 & 3.8 & 0.483 $\pm$ 0.062 & 1.000 $\pm$ 0.000 & 48.3 \\
& NPO & 0.296  $\pm$ 0.036 & 0.997 $\pm$ 0.005 & 29.6 & 0.492 $\pm$ 0.022 & 0.995 $\pm$ 0.000 & 49.2 & 0.414 $\pm$ 0.113 & 0.980 $\pm$ 0.018 & 41.5 \\
\midrule
\multirow{5}{*}{\shortstack[l]{\textbf{William} \\ \textbf{Shakespeare}}} 
& GA  & 0.582 $\pm$ 0.095 & 0.995 $\pm$ 0.009 & 58.2 & 0.283 $\pm$ 0.036 & 0.987 $\pm$ 0.002 & 28.4 & 0.425 $\pm$ 0.013 & 	0.955 $\pm$ 0.005 & 42.7\\
& GD  & 0.663 $\pm$ 0.065 & 1.000 $\pm$ 0.000 & 66.3 & 0.433 $\pm$ 0.159 & 0.995 $\pm$ 0.004 & 43.3 & 0.412 $\pm$ 0.144 & 0.910 $\pm$ 0.072 & 42.2 \\
& KL  & 0.375 $\pm$ 0.066 & 0.597 $\pm$ 0.075 & 55.0 & 0.786 $\pm$ 0.081 & 1.000 $\pm$ 0.000 & 78.6 & 0.558 $\pm$ 0.208 & 0.841 $\pm$ 0.144 & 58.1 \\
& DPO & 0.337 $\pm$ 0.061 & 1.000 $\pm$ 0.000 & 33.7 & 0.050 $\pm$ 0.041 & 0.997 $\pm$ 0.005 & 5.0 & 0.030 $\pm$ 0.030 & 0.820 $\pm$ 0.015 & 18.2 \\
& NPO & 0.245 $\pm$ 0.038 & 1.000 $\pm$ 0.000 & 24.5& 0.417 $\pm$ 0.067 & 0.992 $\pm$ 0.008 & 41.7 &  0.398  $\pm$ 0.140 & 0.941 $\pm$ 0.024 & 40.2\\
\bottomrule
\end{tabular}
}
\end{table*}


The \data{} pipeline were designed to construct fully synthetic datasets for studying LLM unlearning in a controlled environment, mitigating confounding risks with data from the pre-training corpus. In this section, we show how we can make slight tweaks on it to investigate the effects of data inter-connectivity on unlearning data that has been incorporated into the pre-trained model.

Given the vast number of tokens used in training modern LLMs (e.g., 2T for Llama2-7B \citep{llama} and significantly more for advanced versions \citep{dubey2024llama}), it is practically impossible to determine the exact data included or its distribution within the pre-training corpus \citep{shi2023detecting, longpre2023pretrainer}. As a proxy, we select two distinct but widely recognized entities -- `Goldman Sachs' and `William Shakespeare' -- to replace an entity’s name and address within the dense sub-graph of Dataset 2. These entities are prevalent enough to have been included in the pre-training dataset, with `Goldman Sachs' appearing in $1.5$x more web search results than `Willian Shakespeare', a notable consideration given that web pages serve as a primary source for pre-training data. \citep{penedo2024fineweb, wenzek2019ccnet}.

Comparison between the performance of unlearning pre-training data in Table \ref{tab:GSWS_rouge1} and synthetic data in the dense sub-graph in Table \ref{tab:takeaway3} demonstrates that unlearning data presented in the pre-training dataset is indeed more challenging -- as entities in the pre-training data have other relationships embedded within pre-training data, making them more inter-connected compared to the synthetic entities created in Dataset 2. 

The results in Table \ref{tab:GSWS_rouge1} further shows that `Goldman Sachs', as a more inter-connected data point, is more difficult to unlearn compared to `William Shakespeare', evidenced by consistently higher DS scores. These results further support our finding that more inter-connected data are harder to forget. They also highlight the real-world significance of developing and evaluating unlearning methods' capability of addressing the inherent challenges associated with unlearning highly inter-connected data in practical applications.

\section{Related Works}\label{sec:motivation}
\vspace{-0.5em}

%
\textbf{Machine Unlearning. } MU emerged as a tool to exercise the right to be forgotten (RTBF) \citep{european_union_gdpr_2016}. Soon after its inception \citep{cao2015towards}, probabilistic definitions of MU, inspired by differential privacy, were introduced \citep{ginart2019making, sekhari2021remember}. These definitions enabled the development of several certified unlearning algorithms, particularly for deep learning models \citep{guo2019certified, zhang2022prompt, golatkar2020eternal, golatkar2020forgetting}. However, due to efficiency concerns and the difficulty of maintaining strong assumptions, uncertified approximate MU methods have gained more traction \citep{kurmanji2023towards, foster2024fast, sendera2025semu, dang2025efficient}. Early research primarily focused on unstructured data, such as images, with only a few studies exploring structured tabular data \citep{kurmanji2024machine, oesterling2024fair}. 

\textbf{LLM Unlearning.} Recent works have extensively investigated unlearning in generative models, including text-to-image models \citep{fuchi2024erasing, kumari2023ablating, zhang2025defensive} and large language models (LLMs) \citep{shen2025lunar, liu2024rethinking}. The motivations for LLM unlearning extend beyond RTBF and include removing copyrighted content \citep{eldan2023s}, mitigating undesirable behaviors \citep{dige2024can}, and enhancing safety \citep{zhao2024llmunlearningbackdoor}. Existing LLM unlearning methods include simple gradient ascent (GA) based approaches \citep{liu2024rethinking, barbulescu2024each}, preference-optimization-based techniques \citep{rafailov2024direct, zhang2024negative}, surgical activation engineering \citep{shen2025lunar, seyitouglu2024extracting}, and even in-context unlearning \citep{pawelczyk2023context}. 

\textbf{Evaluation and Benchmarks. } Evaluating unlearning methods remains a significant challenge \citep{triantafillou2024arewemaking}, primarily due to two factors: (1) constructing suitable datasets and (2) defining effective evaluation metrics. Unlearning methods are typically assessed using task-specific metrics \cite{kurmanji2023towards} and privacy attacks \cite{triantafillou2024arewemaking}, while some works try to devise new evaluation criteria \cite{shi2024muse}.
From a dataset perspective, researchers often repurpose existing ML datasets for unlearning by introducing a new split, where a subset of training data is designated as the `forget set'. More recently, specialized datasets have been curated for LLM unlearning \cite{maini2024tofu, li2403wmdp}. However, these datasets are usually static, limiting flexibility in size. Moreover, they are susceptible to leakage, potentially influencing the pre-training of newer models and introducing confounding factors when evaluating baselines. A major limitation of existing benchmarks is that they primarily focus on independent data points without accounting for structural dependencies between them. \textbf{\data{} resolves all these problems.}  

\section{Conclusion} \label{sec:roadmap}
\vspace{-0.5em}
In this work, we introduced \data{}, a pipeline for synthesizing structured datasets that uniquely captures the structural relationships among entities to address a critical gap in LLM unlearning: the neglect of data inter-connectivity. Unlike existing benchmarks such as TOFU, \data{} provides clearer evaluation framework and mitigates confounding risks introduced by pre-trained models. Through experiments on datasets synthesized using \data{}, we demonstrated that data inter-connectivity significantly impacts unlearning difficulty. Specifically, we found that (a) unlearning becomes harder as data inter-connectivity grows, (b) the density of the underlying knowledge graph positively correlates with unlearning difficulty, and (c) unlearning domain-skewed data often leads to disproportionate degradation in the retained model’s performance on that same domain.

\section{Limitations} \label{sec:limitations}
While \data{} provides a novel tool for structural dataset compilation for realistic LLM unlearning, several exciting avenues remain open for exploration. First, \data{} currently produces synthetic datasets, which, while carefully designed, may differ from real-world distributions. The studied baselines could behave differently under different distributions. It would be interesting to extend \data{} to compile even naturally structured texts to produce real-world datasets. Second, graph-based representations of structured knowledge exist in multiple modalities, including vision and multimodal learning. Extending our approach beyond textual data could help assess how unlearning manifests in different input formats and model architectures. Finally, our work primarily contributes a dataset compilation pipeline, but a complementary challenge lies in defining robust evaluation metrics for unlearning effectiveness. Developing principled measures that account for several important aspects including forgetting quality, privacy, and model's utility  remains an important direction for future research.

\newpage
\appendix
\onecolumn
\section*{Appendix}

\section{Demand for LLM Unlearning from a Regulatory Perspective}\label{sec:evalmetricsdetails}
The demand for data unlearning is rising due to increased scrutiny over transparency in data usage for training LLMs and concerns regarding the rights of developers to access and use such data. For instance, GDPR grants individuals the right to access all information held by service providers, including details on how data is used for ML training (Art.15, Rec.63 \& 64). Similarly, the European Union’s Artificial Intelligence Act (the EU AI Act), which represents the first comprehensive legal framework of its kind, mandates that model providers publish a comprehensive summary of training data content (Art.52c). 

Additionally, in certain jurisdictions such as the EU, individuals or organizations are legally empowered to demand the deletion of data or withdraw consent for the use of their personal data (\textit{`right-to-be-forgotten'}). A pivotal case that illustrates this is \textit{Google Spain SL v. Agencia Española de Protección de Datos}, where the Court of Justice of the European Union (CJEU) upheld the individual’s right to have links to their personal data removed under specific conditions, even when the original online publication was legal \citep{GooglevAgencia}. The EU AI Act also re-confirmed EU's regulatory stance on clients' right to revoke their consent at any time (Art.17) \citep{woisetschlager2024federated}. 

In the US, legal cases like \textit{"Times v OpenAI"} spotlight the debate over copyright laws' applicability to AI training, leading to broader discussions about prompting legislative measures such as the Generative AI Copyright Disclosure Bill in the US House of Representatives \citep{GenAICopyrightDisclosureAct} to enhance transparency and accountability. These regulatory developments emphasize the importance of effective data erasure practices in ensuring LLMs' legal compliance.

\section{\data{} Contractual Data and QA Template} \label{app:example}

\begin{mybox}{Sales of Goods Contract Template}
\setlist[itemize]{leftmargin=*,label=\textbullet}

\begin{center}
\titlestyle{SALES OF GOODS CONTRACT}
\end{center}
\vspace{2mm}
\subsection*{\underline{1. PARTIES}}
\vspace{2mm}
\begin{itemize}
    \item This Sales Contract (hereinafter referred to as the “\textbf{Contract}”) is entered into on [•] (the “\textbf{Effective Date}”) by and between [•] with an address of [•] (the “\textbf{Seller}”) and [•] with an address of [•] (the “\textbf{Customer}”) (collectively referred to as the “\textbf{Parties}”).
\end{itemize}
\vspace{2mm}

\subsection*{\underline{2. GOODS AND PRICE}}
\vspace{2mm}
\begin{itemize}
    \item The goods that the Seller is selling to the Customer are enlisted below with their quantities (hereinafter referred to as the “\textbf{Goods}”).
    \begin{itemize}
        \item Goods: [•]
        \item Quantity: [•]
        \item Price per unit: [•]
        \item Total price: [•]
    \end{itemize}
\end{itemize}
\vspace{2mm}

\subsection*{\underline{3. PAYMENTS}}
\begin{itemize}
    \item The Seller shall provide the Customer with an invoice no later than [•] days after the time of the delivery.
    \item All invoices are to be paid in full within [•] days. Any balances not paid within [•] days will be subject to a [•]\% late payment penalty. 
\end{itemize}
\vspace{2mm}

\subsection*{\underline{4. DELIVERY AND SHIPPING}}
\begin{itemize}
    \item The delivery of the goods (the “\textbf{Delivery}”) will be at the location [•]. 
    \item The shipping method will be decided by the [•]. [•] will be responsible for the costs of the shipment.
\end{itemize}
\vspace{1.5mm}

\subsection*{\underline{5. WARRANTIES}}
\vspace{1mm}
\begin{itemize}
    \item \textbf{General Warranty:} The Seller hereby warrants to the Customer that the Goods shall be free from defects in materials and workmanship under normal use and service for a period of [•] years from the date of delivery (the "\textbf{Warranty Period}"). The Seller affirms that it has good title to the Goods free and clear of any liens and encumbrances and has the right to sell the Goods to the Customer.
    \item \textbf{Remedy for Breach of Warranty:} In the event of a breach of this warranty, the Customer must notify the Seller in writing within of [•] days of discovering the defect. Upon receiving such notification, the Seller shall, at its sole option, (i) repair or replace the defective Goods at no additional charge to the Customer, or (ii) refund the purchase price paid for the defective Goods, provided that the Goods are returned to the Seller, if so requested. The choice of remedy shall be at the Customer's discretion if repair or replacement does not remedy the defect within a reasonable time.
    \item \textbf{Exclusions from Warranty:} This warranty does not apply to any damage or defect resulting from misuse, abuse, neglect, alterations, unauthorized repairs, modifications, accidents, or natural wear and tear. The Seller's obligation under this warranty is limited to the repair, replacement, or refund as provided under this section and does not cover any other costs such as the cost of removal and reinstallation of Goods, loss of use, loss of profit, or other incidental or consequential damages. 
\end{itemize}

\begin{itemize}

    \item \textbf{No Other Warranties:} Except for the warranty set forth herein, the Seller disclaims all other warranties, express or implied, including, but not limited to, any implied warranties of merchantability or fitness for a particular purpose. The Seller's liability under this warranty shall be limited to the repair, replacement, or refund as specified herein, and in no event shall exceed the purchase price of the defective Goods. 


    \item \textbf{Survival:} This warranty shall survive the delivery, inspection, acceptance, and payment of and for the Goods and shall inure to the benefit of the Customer and its successors and assigns. 
\end{itemize}
\vspace{2mm}

\subsection*{\underline{6. INSPECTION}}
\begin{itemize}
    \item Hereby, the Customer acknowledges that it has relied solely on the investigations, examinations, and inspections that the Customer has chosen to make and that the Seller has afforded the Customer the opportunity for full and complete investigations, examinations, and inspections.
\end{itemize}
\vspace{2mm}

\subsection*{\underline{7. RISK OF LOSS AND TITLE}}
\begin{itemize}
    \item The risk of loss or damage for the goods will be on the Seller until the goods pass upon delivery to the Customer or its designee. The Title of the goods will also remain with the Seller until the goods pass upon delivery to the Customer or its designee.
\end{itemize}
\vspace{2mm}

\subsection*{\underline{8. DELAY OR FAILURE TO PERFORM AND FORCE MAJEURE}}
\begin{itemize}
    \item Under no circumstances will the Seller be held liable to the Customer for any delay that may occur, non-delivery or an arising fault of this Agreement that may be due to any labour dispute, shortage in transportation, delay or shortage of materials to produce the Goods, fires, accidents, Acts of God, or any other causes outside Seller’s control. The Seller will notify the Customer immediately upon realization that it will not be able to deliver the Goods as promised. Upon such notice, either Party may terminate this Agreement.
\end{itemize}
\vspace{2mm}

\subsection*{\underline{9. COOLING-OFF PERIOD}}
\begin{itemize}
    \item Either Party may terminate this Agreement, for any reason, within [•] days following the Effective Date of this Agreement ('\textbf{Cooling-Off Period}'). Termination during this Cooling-Off Period must be communicated in writing to the other Party. Following the expiration of the Cooling-Off Period, no Party shall have the right to terminate this Agreement on the basis of the Cooling-Off Period provisions. 
\end{itemize}
\vspace{2mm}

\subsection*{\underline{10. LIMITATION OF LIABILITY}}
\begin{itemize}
    \item Under no circumstances will the Seller be liable for any indirect, special, consequential, or punitive damages (including lost profits) arising out of or relating to this Agreement or the transactions it contemplates (whether for breach of contract, tort, negligence, or other form of action).
\end{itemize}
\vspace{2mm}

\subsection*{\underline{11. SEVERABILITY}}
\begin{itemize}
    \item In the event that any provision of this Agreement is found to be void and unenforceable by a court of competent jurisdiction, then the remaining provisions will remain in force in accordance with the Parties’ intention.
\end{itemize}
\vspace{2mm}

\subsection*{\underline{12. ENTIRE AGREEMENT}}
\begin{itemize}
    \item This Agreement contains the entire agreement and understanding among the Parties hereto with respect to the subject matter hereof, and supersedes all prior agreements, understandings, inducements and conditions, express or implied, oral or written, of any nature whatsoever with respect to the subject matter hereof. The express terms hereof control and supersede any course of performance and/or usage of the trade inconsistent with any of the terms hereof.
\end{itemize}
\vspace{2mm}

\subsection*{\underline{13. GOVERNING LAW}}
\begin{itemize}
    \item This Agreement shall be governed by and construed in accordance with the laws of [•].
\end{itemize}
\vspace{2mm}

\textbf{The Parties hereby agree to the terms and conditions set forth in this Agreement and such is demonstrated throughout their signatures below.}

\end{mybox}

\begin{mybox}{QAs of the Sales of Goods Contract}
\setlist[itemize]{leftmargin=*,label=\textbullet}
\begin{itemize}
\item Q1: What was the effective date of the contract between [\textit{seller name}] and [\textit{customer name}]?
\vspace{1mm}

\item Q2: What was the name of the seller in the contract with [\textit{customer name}] as of [\textit{effective date}]?
\vspace{1mm}

\item Q3: What was the address of [\textit{seller name}] in the contract with [\textit{customer name}]?
\vspace{1mm}

\item Q4: What was the name of the customer in the contract with [\textit{seller name}] as of [\textit{effective date}]?
\vspace{1mm}

\item Q5: What was the address of [\textit{customer name}] in the contract with [\textit{seller name}]?
\vspace{1mm}

\item Q6: What was the good that the seller was selling to the customer based on the contract between [\textit{seller name}] and [\textit{customer name}]?
\vspace{1mm}

\item Q7:What was the quantity of the good being sold based on the contract between [\textit{seller name}] and [\textit{customer name}]?
\vspace{1mm}

\item Q8: What was the unit price in dollars of the good being sold based on the contract between [\textit{seller name}] and [\textit{customer name}]?
\vspace{1mm}

\item Q9: What was the total price in dollars of the good being sold based on the contract between [\textit{seller name}] and [\textit{customer name}]?
\vspace{1mm}

\item Q10: By how many days after the delivery time must the seller provide the customer with an invoice based on the contract between [\textit{seller name}] and [\textit{customer name}]?
\vspace{1mm}

\item Q11: Within how many days must the invoice be paid in full based on the contract between [\textit{seller name}] and [\textit{customer name}]?
\vspace{1mm}

\item Q12: After how many days would unpaid balances incur a late payment penalty based on the contract between [\textit{seller name}] and [\textit{customer name}]?
\vspace{1mm}

\item Q13: What was the late payment interest rate based on the contract between [\textit{seller name}] and [\textit{customer name}]?
\vspace{1mm}

\item Q14: What was the address of delivery based on the contract between [\textit{seller name}] and [\textit{customer name}]?
\vspace{1mm}

\item Q15: Who would decide the shipping method based on the contract between [\textit{seller name}] and [\textit{customer name}]?
\vspace{1mm}

\item Q16: Who would be responsible for the costs of the shipment based on the contract between [\textit{seller name}] and [\textit{customer name}]?
\vspace{1mm}

\item Q17: What was the duration of the general warranty period in years based on the contract between [\textit{seller name}] and [\textit{customer name}]?
\vspace{1mm}

\item Q18: Within how many days of discovering a defect must the customer notify the seller in writing in the event of a breach of warranty based on the contract between [\textit{seller name}] and [\textit{customer name}]?
\vspace{1mm}

\item Q19: What was the duration of the cooling-off period in days based on the contract between [\textit{seller name}] and [\textit{customer name}]?
\vspace{1mm}

\item Q20: Which jurisdiction's laws govern the contract between [\textit{seller name}] and [\textit{customer name}]?
\end{itemize}
\end{mybox}

\begin{mybox}{Employment Contract Template}
\setlist[itemize]{leftmargin=*,label=\textbullet}

\begin{center}
\titlestyle{EMPLOYMENT CONTRACT}
\end{center}

\subsection*{\underline{1. PARTIES}}
\begin{itemize}
    \item [•] (the "\textbf{Employer}") with its principal place of business located at [•] ("\textbf{Employer’s Business Address"}) agrees to employ [•] (the "\textbf{Employee}") who lives at [•] ("\textbf{Employee’s Residential Address}") and the Employee agrees to be employed, on the terms and conditions set out in this Contract, and in the accompanying Addendum (together, the "\textbf{Agreement}").
\end{itemize}
\vspace{2mm}

\subsection*{\underline{2.	START AND LENGTH OF EMPLOYMENT}}
\begin{itemize}
    \item The Employee will start employment on [•] ("\textbf{Start Date}"). 
    \item The Employer shall employ the Employee for [•] months ("\textbf{Length of Employment}"), however, the Employer and the Employee may change the Length of Employment in accordance with Clause 12 of this Contract.
\end{itemize}
\vspace{2mm}

\subsection*{\underline{3.	JOB TITLE AND DUTIES}}
\begin{itemize}
    \item The Employee shall be employed as [•] ("\textbf{Position}"). The Employee shall perform the duties as described in the accompanying Addendum, and any other duties reasonably assigned by the Employer. 
\end{itemize}
\vspace{2mm}

\subsection*{\underline{4.	PLACE OF WORK}}
\begin{itemize}
    \item The Employee shall work at [•] ("\textbf{Address of Work Location}"). The Employee shall not be required to work at a different location unless the Employee consents in writing to such an arrangement. Any such employment shall be on the same terms and conditions as this Agreement.
\end{itemize}
\vspace{2mm}

\subsection*{\underline{5.	WORKING HOURS}}
\begin{itemize}
    \item The Employee’s normal days of work are Monday to Friday ("\textbf{Normal Work Days}") and the Employee’s normal hours of work are [•] to [•] ("\textbf{Normal Work Hours}") (together, the "\textbf{Work Week}"). 
\end{itemize}
\vspace{2mm}

\subsection*{\underline{6. PAY}}
\begin{itemize}
    \item The Employer shall pay the Employee \$[•] ("\textbf{Rate of Basic Pay}") per hour. 
    \item The Employer shall pay the Employee in [•] instalments.
\end{itemize}
\vspace{2mm}

\subsection*{\underline{7. BENEFITS}}
\begin{itemize}
    \item The Employee shall be entitled to participate in [•] offered by the Employer, subject to the terms and conditions of those plans.
\end{itemize}
\vspace{2mm}

\subsection*{\underline{8.	HOLIDAYS}}
\begin{itemize}
    \item The Employer shall provide the Employee with [•] days of paid holiday leave ("\textbf{Holiday Leave}") per year, plus the public holidays. 
    \item The Employee shall provide the Employer two weeks’ notice of any Holiday Leave, and the Employer may only refuse Holiday Leave in exceptional circumstances. The Employer shall pay the Employee for any unused Holiday Leave at the earlier of: (i) the end of each year, or (ii) the end of the Employee’s employment.
\end{itemize}
\vspace{2mm}

\subsection*{\underline{9.	CONFIDENTIALITY}}
\begin{itemize}
    \item The Employee agrees that during the term of employment and for the first [•] months thereafter, he/she will not disclose any confidential information pertaining to the business of the Employer to any person not authorized by the Employer to receive such information.
\end{itemize}
\vspace{2mm}



\subsection*{\underline{10. WORK CONDITIONS}}
\begin{itemize}
    \item The Employer shall ensure the Employee is appropriately instructed and trained in relation to tasks that the Employee will carry out. The Employer shall provide a safe and healthy work environment and shall not require the Employee to do work that subjects the Employee to health or safety hazards.
\end{itemize}
\vspace{2mm}

\subsection*{\underline{11. SICK PAY AND ABSENCE}}
\begin{itemize}
    \item The Employee shall notify the Employer if he or she is going to be absent from work because of sickness or injury. The Employer shall not require the Employee to work when sick or injured. 
    \item In each year of employment, the Employee shall be entitled to receive the Basic Rate of Pay (as if he/she had worked the Normal Work Hours) per day for the first [•] days of absence from work due to sickness or injury ("\textbf{Paid Sick Leave}"). 
\end{itemize}
\vspace{2mm}

\subsection*{\underline{12.	TERMINATION}}
\begin{itemize}
    \item The Employee and Employer shall each provide the other with [•] weeks’ written notice of termination in a language the Employee understands. 
    \item The Employee and Employer may agree that the Employer pay the Employee for this notice period instead of requiring the Employee to work. In exceptional circumstances, as defined in the Addendum, notice of termination is not required.
    \item On termination, the Employee shall return to the Employer all Employer property, and the Employer shall pay immediately all monies due under this Agreement to the Employee.
\end{itemize}
\vspace{2mm}

\subsection*{\underline{13.	NON-COMPETE}}
\begin{itemize}
    \item During the term of employment and for [•] months after the termination of employment, the Employee agrees not to engage in any business activities or employment with a competitor or in any capacity that directly competes with the Employer's business within the United States. 
    \item This restriction applies to similar products, services, or industry sectors in which the Employer operates. The Employee acknowledges that such competition could harm the Employer's business interests and agrees to refrain from such activities to protect the Employer's legitimate business interests.

\end{itemize}
\vspace{2mm}

\subsection*{\underline{14.	CHANGES TO EMPLOYMENT TERMS}}
\begin{itemize}
    \item This Contract and the attached Addendum make up the entire Agreement relating to the Employee’s employment. The Employer shall not make any changes to this Agreement without the Employee’s written consent. The Employer shall provide [•] weeks’ written notice of any proposed changes in a language the Employee understands, and the Employer shall permit the Employee to ask questions about such changes. 
\end{itemize}
\vspace{2mm}

\subsection*{\underline{15.	ENTIRE AGREEMENT}}
\begin{itemize}
    \item This Agreement and the attached Addendum contain the entire agreement between the parties. The Employee acknowledges that he/she has not relied on any oral or written representations made by the Employer or its employees or agents.
\end{itemize}
\vspace{2mm}

\subsection*{\underline{16.	GOVERNING LAW}}
\begin{itemize}
    \item This Agreement and any dispute or claim arising out of or in connection with it or its subject matter or formation (including non-contractual disputes or claims) shall be governed by and construed in accordance with the laws of [•] ("\textbf{Governing Law}").
\end{itemize}
\vspace{2mm}

\textbf{I acknowledge that I have read this Contract and the Addendum to this Contract; I understand and accept the terms and conditions set out within it, and that this Contract, together with the Addendum, form the Agreement of Employment.}

\end{mybox}

\begin{mybox}{QAs of the Employment Contract}
\setlist[itemize]{leftmargin=*,label=\textbullet}
\begin{itemize}
\item Q1: What was the name of the employer in the employment contract with [\textit{employee name}], which started from [\textit{start date}]?
\vspace{1mm}

\item Q2: What was the principal business location of [\textit{employer name}] based on the contract between [\textit{employer name}] and [\textit{employee name}]?
\vspace{1mm}

\item Q3: What was the name of the employee in the employment contract with [\textit{employer name}], which started from [\textit{start date}]?
\vspace{1mm}

\item Q4: What was the address of [\textit{employee name}] based on the contract between [\textit{employer name}] and [\textit{employee name}]?
\vspace{1mm}

\item Q5: What was the start date based on the contract between [\textit{employer name}] and [\textit{employee name}]?
\vspace{1mm}

\item Q6: For how many months will the employer employ the employee based on the contract between [\textit{employer name}] and [\textit{employee name}]?
\vspace{1mm}

\item Q7: What was the job position based on the contract between [\textit{employer name}] and [\textit{employee name}]? 
\vspace{1mm}

\item Q8: What was the work location based on the contract between [\textit{employer name}] and [\textit{employee name}]?
\vspace{1mm}

\item Q9: At what hour did the workday start based on the contract between [\textit{employer name}] and [\textit{employee name}]?
\vspace{1mm}

\item Q10: At what hour did the workday finish based on the contract between [\textit{employer name}] and [\textit{employee name}]?
\vspace{1mm}

\item Q11: What was the hourly basic pay in dollars based on the contract between [\textit{employer name}] and [\textit{employee name}]?
\vspace{1mm}

\item Q12: What was the frequency of salary payment based on the contract between [\textit{employer name}] and [\textit{employee name}]?
\vspace{1mm}

\item Q13: What benefit was provided to the employee based on the contract between [\textit{employer name}] and [\textit{employee name}]? 
\vspace{1mm}

\item Q14: How many days of paid holiday leave were provided to the employee based on the contract between [\textit{employer name}] and [\textit{employee name}]?
\vspace{1mm}

\item Q15: For how many months after the employment ends was the employee prohibited from disclosing any confidential information based on the contract between [\textit{employer name}] and [\textit{employee name}]? 
\vspace{1mm}

\item Q16: What was the number of days the employee was entitled to Paid Sick Leave in each year of employment based on the contract between [\textit{employer name}] and [\textit{employee name}]? 
\vspace{1mm}

\item Q17: How many weeks' written notice of termination must the employee and employer each provide to the other based on the contract between [\textit{employer name}] and [\textit{employee name}]?
\vspace{1mm}

\item Q18: For how many months did the non-compete clause cover based on the contract between [\textit{employer name}] and [\textit{employee name}]? 
\vspace{1mm}

\item Q19: How many weeks' written notice must the employer provide before any proposed changes to the terms of employment based on the contract between [\textit{employer name}] and [\textit{employee name}]?
\vspace{1mm}

\item Q20: Which jurisdiction's laws govern the contract between [\textit{employer name}] and [\textit{employee name}]?

\end{itemize}
\end{mybox}

\section{Details of Dataset Construction under \data{}}\label{app:sample_dataset}
Each dataset is organized into columns of questions, answers, and edges, to facilitate easier selection of unlearning edges (i.e., unlearning data). The edge features in Dataset 1 consisted of the placeholder name, such as $AC$. In Dataset 2, each inter-connected sub-component consists of $10$ nodes. The nodes are sequentially numbered: $0-9$ for the sparse sub-graph, $10-19$ for the semi-dense sub-graph, and $20-29$ for the dense sub-graph. The sparse sub-component has a chain structure, with edges sequentially connecting nodes from $0$ to $9$. The semi-dense sub-component contains $27$ edges. The dense sub-component is a fully-connected sub-graph, meaning every pair of nodes within the sub-graph is linked by an edge. 
%
%
The dataset can be found in Hugging Face \url{https://huggingface.co/datasets/xinchiqiu/PISTOL}.

\paragraph{Code of Ethics:}
The \data{} dataset creation pipeline as well as Dataset 1 and Dataset 2 are constructed in the manuscript do not involve any human subjects or participants. Comprehensive documentation will be maintained alongside the dataset, detailing its structure, the nature of the data, and instructions for its use. This documentation will be updated with each version of the dataset.
As mentioned in the manuscript, we set the contract template each with 20 attributes to be filled in. We focused on two ubiquitous types of contracts, sales of goods and employment contracts, owing to their more standardized structure in contrast to other highly customized agreements like corporate sale and purchase agreements or share subscription agreements. Also, we generate attributes in a random manner, taking into account the dataset size. In our datasets, we randomly generate 6 letters and a suffix for a company name (e.g. Empblq LLC), 4 letters for the first name and the surname of a person (e.g. Jkeq Cyfz), 3 numbers, 6 letters, and a street type for an address (e.g. 442 Rcvvyy Boulevard). Other attributes such as the signing date, contractual terms, and governing jurisdiction are also randomly generated. 
Therefore, there are no privacy and copyright implications associated with our datasets. Please note that the contracts used in sample datasets are generated in a completely random manner, hence do not represent any real contracts between any real companies or individuals.

\section{Evaluation Metrics}\label{app:evalmetricsdetails}
The evaluation of unlearning presents significant challenges. ~\cite{thudi2022necessity} demonstrates that, in certain scenarios, it is impossible to audit unlearning processes using the single metric of model losses 
even with access to the entire training trajectory. 
Although this underscores the inherent difficulties of unlearning evaluations, the analysis in \cite{thudi2022necessity} does not preclude using other heuristic-based methods to assess unlearning.  
To address this, we propose to employ multiple metrics. 
The use of multiple diverse metrics allows us to alleviate the unlearning evaluation trap that certain data points of a equivalent class would produce the same metric change without effective target  removal~\cite{thudi2022necessity}. Prior unlearning benchmark use ROUGE~\cite{maini2024tofu}. Given our focus on \textit{structural} LLMs unlearning,  we also incorporate metrics like MRR and hit ratios as they are representative metrics for structured learning communities. 
ROUGE score~\cite{lin-hovy-2003-automatic,lin-och-2004-automatic,lin2004rouge} is commonly used for text-generation tasks (e.g., QA tasks), while MRR and hit ratios are popular for entity retrieval-type tasks (e.g., knowledge graph completion)~\cite{pmlr-v48-trouillon16,lacroix2018canonical,chen2021relation}. Given our focus on benchmarking structural LLMs unlearning, we incorporate metrics from both of these communities to provide a comprehensive evaluation.

\paragraph{ROUGE score:} We use ROUGE scores to compare model answers (with greedy sampling) with the ground truth. Specifically, we compute the ROUGE-1 recall score \cite{lin2004rouge}, which acts as a surrogate for accuracy on the question-answering task, as it accounts for the output phrasing to be slightly different than the ground truth.

\paragraph{Mean reciprocal rank (MRR).} An answer is usually composed of multiple tokens. Therefore, we use the reciprocal average of the rank of each target (ground truth) token to measure the model’s memorization of names. Given a prefix $Q$, an output answer token sequence $E = {e_1, ..., e_n}$, with the length of $|E|$, the model predicts the rank of the target token as
$rank(e_i|Q)$, and then MRR for the name $E$ is calculated as follows:
\begin{equation}
    MRR = \frac{\sum_{i=1}^{|E|} 1/rank(e_i,Q)}{|E|}
\end{equation}

\paragraph{Top hit ratio (THR)} The hit rate is a binary score for each output token, indicating the presence of the correct token at the top $m$ values in the output logits, denotes as $hit(e_i, m)$. Also, given the output sequence $E = {e_1, ..., e_n}$, and we choose $m=100$ in our experiments.
\begin{equation}
    THR = \frac{\sum_{i=1}^{|E|}hit(e_i, m)}{|E|}
\end{equation}


\section{Model Fine-tuning}\label{sec:finetuneing}
Datasets constructed under \data{} are synthetic with structured Q\&As derived from randomly generated contractual attributes. As such, pre-trained model must first be fine-tuned on the constructed dataset to ensure the model effectively `remembers' the new data points. After fine-tuning, we inference the model on questions of our dataset, and recorded that all the questions can be successfully learned and remembered by both models. As discussed in Section \ref{sec:contractdataset}, fine-tuning on datasets constructed under \data{}, by its design, facilitates more accurate evaluation on $\mathcal{D}_{fact}$ later on due to elimination of confounding variables.

For fine-tuning, we implemented the widely adopted parameter-efficient fine-tuning method LoRA \cite{lora} in our experiments. LoRA saves computation memory by optimizing over two low rank metrics $B,A$, where $BA = \triangle w$, instead of the entire parameters space. In all of our experiments, we optimize this loss with AdamW for $20$ epochs and warm up for the first epoch. We use an effective batch size of $16$. We verify that the LLM can accurately `remember' all individual data points and reaches ROUGE1 score of 1 for all models. It is worth noting that Dataset 1 is designed to be both concise and effective for studying structural LLM unlearning. This design ensures accessibility for researchers with limited computational resources. For those with larger computational budgets, Dataset 2 can be utilized, or larger datasets can be constructed in a similar manner using the \data{} pipeline. Finetuning can be run using 1 NVIDIA A40 GPU, and the running times depend on the model and the
size of dataset up to 2 hours.


\section{Machine Unlearning}\label{sec:unlearningmethod}


We experiment with several unlearning methods summarised in the survey paper \cite{liu2024rethinking, maini2024tofu}, each of which is introduced in detail in the section. Given the nascent state of LLM unlearning and the general lack of robustness in existing unlearning methods, we select mainstream approaches from two major families -- gradient ascent-based and preference optimization-based methods -- to illustrate the impact of structured data on LLM unlearning performance.

\paragraph{Gradient Ascent (GA).} GA is the most straightforward and intuitive method, performing gradient ascent on the forget data to maximize the likelihood of mispredictions for those samples within the forget set $\mathcal{D}_f$ \cite{jang2022knowledge, yao2023large}, according to the loss function:
\begin{equation}
    \mathcal{L}_{\phi}(\mathcal{D}_f) = \frac{1}{|\mathcal{D}_{f}|}\sum_{x\in \mathcal{D}_{f}} l_{\phi}(x)
\end{equation}
It is worth noting that GA alone can be sensitive to the choice of hyperparameters during optimization, such as the number of ascent steps and the learning rate. Therefore, during the unlearning stage, the loss we aim to maximize is the average over the forget set $\mathcal{D}_f$.

\paragraph{Gradient Difference (GD).} Grad Difference \cite{liu2022continual} extends the idea of GA by optimizing two losses: one maximizes mispredictions on the forget set and the other minimizes mispredictions on the retained set, thus simultaneously unlearning the forget set and maintaining performance on the retained set. The combined loss function is: 
\begin{equation}
    \mathcal{L}_{\phi} = - \mathcal{L}_{\phi}(\mathcal{D}_f) + \mathcal{L}_{\phi}(\mathcal{D}_r) 
\end{equation}
Given the fact that the size of the forget set is normally smaller than the retained set (otherwise, it will be more computationally efficient to simply retrain on the retained set), the mini-batch selection follows the selection from the forget set first, and then for each selected forget samples, we randomly select a retained sample to form a combined sample for the loss computation.

\paragraph{Unlearning with KL-divergence (UKL).} The UKL method aims to minimize the KL-divergence between the predictions of the original fine-tuned model and the unlearned model on the retained set $\mathcal{D}_r$, thereby maximizing the utility of the model on the retained data, while concurrently maximizing the loss on the forget set \cite{maini2024tofu}. The loss function can be expressed as below:
\begin{equation}
    \mathcal{L}_{\phi} = -\mathcal{L}_{\phi}(\mathcal{D}_f) + \frac{1}{|\mathcal{D}_r|} \sum_{x\in \mathcal{D}_r} \frac{1}{|x|} \sum_{i=2}^{|x|} KL(M_{pretrained}(x_{<i} || M_{unlearn}(x_{<i}))
\end{equation}

\paragraph{Direct Preference Optimization (DPO).} DPO aims to align the model such that it refrains from revealing information from the forget set. The approach, inspired by the original DPO method \cite{rafailov2024direct} and following the TOFU framework \cite{maini2024tofu}, computes the loss using $x_{idk} = [q, a_{idk}]$, which are question-answer pairs from the forget set \(\mathcal{D}_f\) but with the answer replaced by various expressions of 'I don't know'. Unlike prior algorithms, DPO does not utilize gradient ascent. The loss function is:
\begin{equation}
    \mathcal{L}_{\phi} = \mathcal{L}_{\phi}(\mathcal{D}_r) + \mathcal{L}_{\phi}(\mathcal{D}_{f,idk})
\end{equation}

\paragraph{Negative Preference Optimization (NPO).} NPO \cite{zhang2024negative} is effectively a modified version of DPO that excludes positive samples (i.e., $\mathcal{L}_{\phi}(\mathcal{D}_r)$). The gradient of NPO can also be interpreted as an adaptive weighting of the GA gradient, where the weight vanishes for unlearned samples. This mitigates the unbounded nature of simple gradient ascent, which often leads to catastrophic collapse of the unlearned model when hyperparameters are not properly tuned.

\paragraph{Unlearning configurations.} For all unlearning methods, we conduct optimization of the corresponding loss over $20$ epochs. In scenarios where support from the retained set is utilized, an epoch is defined as one complete cycle through the entire forget set, using no more than the same number of samples from the retained set. We employ the AdamW optimizer with a warm-up phase during the first epoch and maintain an effective batch size of $4$ for all unlearning algorithms. We evaluated learning rates between $1\times10^{-6}$ and $5\times10^{-5}$. Since successful unlearning must preserve model utility on the retained dataset, we enforce a performance threshold and select the learning rate that maximizes forgetting. All unlearning experiments can be run using 1 NVIDIA A40 GPU, and the running time depends on the size and the algorithms.


\newpage

\section{Additional Experiment Results} \label{app:additionalres}

In this section, we show additional experimental results, reporting MRR and THR metrics.

\begin{table*}[ht]
\centering
\captionsetup{font=small,labelfont=bf}
\caption{Additional results of unlearning data in Dataset 1 with different degrees of inter-connectivity.}
\label{tab:additional_res_dataset1_interconnectivity}
\scalebox{0.7}{
\begin{tabular}{ll |cc |cc}
\toprule
{\textbf{Forget Data}} & 
{\textbf{Forget Method}} & 
{\textbf{Forget Set MRR}} & 
{\textbf{Retain Set MRR}} & 
{\textbf{Forget Set THR}} & 
{\textbf{Retain Set THR}} \\
\midrule
\multicolumn{6}{l}{\textbf{Llama2-7B}} \\

\multirow{5}{*}{\textbf{AB}} 
& GA  & 0.288 $\pm$ 0.004 & 0.310 $\pm$ 0.004 & 0.766 $\pm$ 0.024 & 0.759 $\pm$ 0.004 \\
& GD  & 0.277 $\pm$ 0.008 & 0.292 $\pm$ 0.002 & 0.731 $\pm$ 0.021 & 0.733 $\pm$ 0.006 \\
& KL  & 0.355 $\pm$ 0.014 & 0.370 $\pm$ 0.017 & 0.839 $\pm$ 0.016 & 0.831 $\pm$ 0.010 \\
& DPO & 0.261 $\pm$ 0.011 & 0.295 $\pm$ 0.002 & 0.600 $\pm$ 0.003 & 0.690 $\pm$ 0.009 \\
& NPO & 0.234 $\pm$ 0.001 & 0.281 $\pm$ 0.003 & 0.710 $\pm$ 0.008 & 0.782 $\pm$ 0.001 \\
\midrule
\multirow{5}{*}{\textbf{AC}} 
& GA  & 0.179 $\pm$ 0.001 & 0.271 $\pm$ 0.004 & 0.606 $\pm$ 0.004 & 0.735 $\pm$ 0.006 \\
& GD  & 0.168 $\pm$ 0.002 & 0.261 $\pm$ 0.002 & 0.590 $\pm$ 0.009 & 0.728 $\pm$ 0.008 \\
& KL  & 0.357 $\pm$ 0.031 & 0.407 $\pm$ 0.044 & 0.694 $\pm$ 0.073 & 0.784 $\pm$ 0.047 \\
& DPO & 0.159 $\pm$ 0.001 & 0.277 $\pm$ 0.001 & 0.425 $\pm$ 0.003 & 0.658 $\pm$ 0.009 \\
& NPO & 0.186 $\pm$ 0.002 & 0.292 $\pm$ 0.001 & 0.651 $\pm$ 0.014 & 0.810 $\pm$ 0.001 \\
\midrule
\multicolumn{6}{l}{\textbf{Mistral-7B}} \\

\multirow{5}{*}{\textbf{AB}} 
& GA  & 0.166 $\pm$ 0.014 & 0.269 $\pm$ 0.004 & 0.500 $\pm$ 0.013 & 0.678 $\pm$ 0.007 \\
& GD  & 0.163 $\pm$ 0.002 & 0.285 $\pm$ 0.004 & 0.524 $\pm$ 0.044 & 0.713 $\pm$ 0.013 \\
& KL  & 0.293 $\pm$ 0.028 & 0.316 $\pm$ 0.008 & 0.737 $\pm$ 0.009 & 0.784 $\pm$ 0.004 \\
& DPO & 0.081 $\pm$ 0.019 & 0.271 $\pm$ 0.004 & 0.187 $\pm$ 0.043 & 0.631 $\pm$ 0.019 \\
& NPO & 0.580 $\pm$ 0.022 & 0.690 $\pm$ 0.019 & 0.838 $\pm$ 0.004 & 0.914 $\pm$ 0.002 \\
\midrule
\multirow{5}{*}{\textbf{AC}} 
& GA  & 0.134 $\pm$ 0.054 & 0.236 $\pm$ 0.004 & 0.252 $\pm$ 0.079 & 0.597 $\pm$ 0.025 \\
& GD  & 0.149 $\pm$ 0.006 & 0.266 $\pm$ 0.005 & 0.293 $\pm$ 0.047 & 0.635 $\pm$ 0.016 \\
& KL  & 0.246 $\pm$ 0.002 & 0.279 $\pm$ 0.008 & 0.727 $\pm$ 0.019 & 0.757 $\pm$ 0.008 \\
& DPO & 0.018 $\pm$ 0.001 & 0.269 $\pm$ 0.004 & 0.125 $\pm$ 0.000 & 0.676 $\pm$ 0.001 \\
& NPO & 0.231 $\pm$ 0.008 & 0.335 $\pm$ 0.005 & 0.616 $\pm$ 0.003 & 0.790 $\pm$ 0.002 \\
\midrule
\multicolumn{6}{l}{\textbf{{Gemma-7B}}} \\

\multirow{5}{*}{\textbf{AB}} 
& GA  & 0.706 $\pm$ 0.013 & 0.916 $\pm$ 0.002 & 0.797 $\pm$ 0.008 & 0.944 $\pm$ 0.006 \\
& GD  & 0.527 $\pm$ 0.072 & 0.888 $\pm$ 0.023 & 0.652 $\pm$ 0.028 & 0.930 $\pm$ 0.006 \\
& KL  & 0.838 $\pm$ 0.022 & 0.943 $\pm$ 0.004 & 0.923 $\pm$ 0.015 & 0.978 $\pm$ 0.003 \\
& DPO & 0.894 $\pm$ 0.010 & 1.000 $\pm$ 0.000 & 0.954 $\pm$ 0.009 & 1.000 $\pm$ 0.000 \\
& NPO & 0.526 $\pm$ 0.047 & 0.897 $\pm$ 0.015 & 0.716 $\pm$ 0.024 & 0.941 $\pm$ 0.008 \\
\midrule
\multirow{5}{*}{\textbf{AC}} 
& GA  & 0.769 $\pm$ 0.014 & 0.954 $\pm$ 0.005 & 0.808 $\pm$ 0.016 & 0.983 $\pm$ 0.007 \\
& GD  & 0.655 $\pm$ 0.010 & 0.947 $\pm$ 0.002 & 0.700 $\pm$ 0.005 & 0.985 $\pm$ 0.002 \\
& KL  & 0.788 $\pm$ 0.047 & 0.960 $\pm$ 0.003 & 0.924 $\pm$ 0.012 & 0.979 $\pm$ 0.003 \\
& DPO & 0.071 $\pm$ 0.003 & 0.739 $\pm$ 0.019 & 0.316 $\pm$ 0.049 & 0.862 $\pm$ 0.010 \\
& NPO & 0.663 $\pm$ 0.022 & 0.939 $\pm$ 0.006 & 0.768 $\pm$ 0.004 & 0.967 $\pm$ 0.007 \\
\bottomrule
\end{tabular}
}
\end{table*}

\begin{table*}[ht]
\centering
\captionsetup{font=small,labelfont=bf}
\caption{Additional results of unlearning data in Dataset 2 with different knowledge densities.}
\label{tab:additional_res_dataset2_densities}
\scalebox{0.7}{
\begin{tabular}{ll |cc |cc}
\toprule
{\textbf{Forget Data}} & 
{\textbf{Forget Method}} & 
{\textbf{Forget Set MRR}} & 
{\textbf{Retain Set MRR}} & 
{\textbf{Forget Set THR}} & 
{\textbf{Retain Set THR}} \\
\midrule
\multicolumn{6}{l}{\textbf{Llama2-7B}} \\

\multirow{5}{*}{\textbf{Dense}} 
& GA  & 0.215 $\pm$ 0.037 & 0.276 $\pm$ 0.001 & 0.573 $\pm$ 0.062 & 0.608 $\pm$ 0.008 \\
& GD  & 0.223 $\pm$ 0.032 & 0.277 $\pm$ 0.001 & 0.583 $\pm$ 0.049 & 0.610 $\pm$ 0.008 \\
& UKL  & 0.268 $\pm$ 0.048 & 0.376 $\pm$ 0.005 & 0.521 $\pm$ 0.149 & 0.593 $\pm$ 0.063 \\
& DPO & 0.219 $\pm$ 0.053 & 0.278 $\pm$ 0.002 & 0.470 $\pm$ 0.100 & 0.602 $\pm$ 0.006 \\
& NPO & 0.136 $\pm$ 0.037 &	0.296 $\pm$ 0.011 &	0.421 $\pm$ 0.052 &	0.576 $\pm$ 0.002 
\\
\midrule
\multirow{5}{*}{\textbf{Semi-Dense}} 
& GA  & 0.197 $\pm$ 0.039 & 0.277 $\pm$ 0.001 & 0.544 $\pm$ 0.058 & 0.609 $\pm$ 0.006 \\
& GD  & 0.240 $\pm$ 0.053 & 0.277 $\pm$ 0.001 & 0.584 $\pm$ 0.086 & 0.609 $\pm$ 0.005 \\
& UKL  & 0.255 $\pm$ 0.073 & 0.333 $\pm$ 0.029 & 0.422 $\pm$ 0.114 & 0.518 $\pm$ 0.086 \\
& DPO & 0.205 $\pm$ 0.042 & 0.279 $\pm$ 0.001 & 0.460 $\pm$ 0.086 & 0.608 $\pm$ 0.010 \\
& NPO & 0.162 $\pm$ 0.042 &	0.293 $\pm$ 0.009 &	0.421 $\pm$ 0.040 &	0.575 $\pm$ 0.005 \\
\midrule
\multirow{5}{*}{\textbf{Sparse}} 
& GA  & 0.179 $\pm$ 0.039 & 0.278 $\pm$ 0.006 & 0.516 $\pm$ 0.056 & 0.610 $\pm$ 0.009 \\
& GD  & 0.202 $\pm$ 0.054 & 0.279 $\pm$ 0.003 & 0.522 $\pm$ 0.052 & 0.608 $\pm$ 0.009 \\
& UKL  & 0.226 $\pm$ 0.045 & 0.366 $\pm$ 0.022 & 0.427 $\pm$ 0.137 & 0.564 $\pm$ 0.061 \\
& DPO & 0.158 $\pm$ 0.040 & 0.280 $\pm$ 0.005 & 0.400 $\pm$ 0.035 & 0.599 $\pm$ 0.019 \\
& NPO & 0.177 $\pm$ 0.020 & 0.273 $\pm$ 0.005 &	0.479 $\pm$ 0.005 & 0.548 $\pm$ 0.002 \\
\midrule
\multicolumn{6}{l}{\textbf{Mistral-7B}} \\

\multirow{5}{*}{\textbf{Dense}} 
& GA  & 0.264 $\pm$ 0.079 & 0.318 $\pm$ 0.015 & 0.566 $\pm$ 0.117 & 0.685 $\pm$ 0.031 \\
& GD  & 0.273 $\pm$ 0.074 & 0.320 $\pm$ 0.013 & 0.588 $\pm$ 0.142 & 0.683 $\pm$ 0.033 \\
& UKL  & 0.318 $\pm$ 0.033 & 0.319 $\pm$ 0.003 & 0.764 $\pm$ 0.038 & 0.757 $\pm$ 0.006 \\
& DPO & 0.039 $\pm$ 0.028 & 0.307 $\pm$ 0.005 & 0.158 $\pm$ 0.095 & 0.667 $\pm$ 0.017 \\
& NPO & 0.164 $\pm$ 0.024 & 0.327 $\pm$ 0.008 & 0.524 $\pm$ 0.000 & 0.736 $\pm$ 0.002 \\
\midrule
\multirow{5}{*}{\textbf{Semi-Dense}} 
& GA  & 0.380 $\pm$ 0.087 & 0.977 $\pm$ 0.012 & 0.540 $\pm$ 0.094 & 0.669 $\pm$ 0.069 \\
& GD  & 0.377 $\pm$ 0.089 & 0.983 $\pm$ 0.007 & 0.550 $\pm$ 0.099 & 0.668 $\pm$ 0.080 \\
& UKL  & 0.803 $\pm$ 0.110 & 0.999 $\pm$ 0.002 & 0.721 $\pm$ 0.045 & 0.751 $\pm$ 0.004 \\
& DPO & 0.075 $\pm$ 0.056 & 0.991 $\pm$ 0.004 & 0.187 $\pm$ 0.068 & 0.671 $\pm$ 0.013 \\
& NPO & 0.151 $\pm$ 0.046 & 0.323 $\pm$ 0.002 & 0.493 $\pm$ 0.066 & 0.716 $\pm$ 0.023 \\
\midrule
\multirow{5}{*}{\textbf{Sparse}} 
& GA  & 0.223 $\pm$ 0.049 & 0.311 $\pm$ 0.006 & 0.517 $\pm$ 0.100 & 0.657 $\pm$ 0.051 \\
& GD  & 0.218 $\pm$ 0.048 & 0.313 $\pm$ 0.008 & 0.520 $\pm$ 0.110 & 0.662 $\pm$ 0.044 \\
& UKL  & 0.290 $\pm$ 0.044 & 0.314 $\pm$ 0.009 & 0.665 $\pm$ 0.152 & 0.735 $\pm$ 0.045 \\
& DPO & 0.028 $\pm$ 0.013 & 0.309 $\pm$ 0.002 & 0.142 $\pm$ 0.025 & 0.670 $\pm$ 0.017 \\
& NPO & 0.186 $\pm$ 0.063 & 0.311 $\pm$ 0.010 & 0.384 $\pm$ 0.040 & 0.665 $\pm$ 0.017 \\
\midrule
\multicolumn{6}{l}{\textbf{Gemma-7B}} \\
\multirow{5}{*}{\textbf{Dense}} 
& GA  & 0.562 $\pm$ 0.108 & 0.972 $\pm$ 0.006 & 0.686 $\pm$ 0.076 & 0.997 $\pm$ 0.002 \\
& GD  & 0.542 $\pm$ 0.101 & 0.984 $\pm$ 0.009 & 0.660 $\pm$ 0.045 & 0.997 $\pm$ 0.002 \\
& UKL  & 0.671 $\pm$ 0.110 & 0.856 $\pm$ 0.062 & 0.764 $\pm$ 0.124 & 0.932 $\pm$ 0.033 \\
& DPO & 0.066 $\pm$ 0.028 & 0.880 $\pm$ 0.014 & 0.211 $\pm$ 0.001 & 0.951 $\pm$ 0.006 \\
& NPO & 0.554 $\pm$ 0.118 & 0.982 $\pm$ 0.006 & 0.723 $\pm$ 0.093 & 0.996 $\pm$ 0.005 \\
\midrule
\multirow{5}{*}{\textbf{Semi-Dense}} 
& GA  & 0.514 $\pm$ 0.120 & 0.888 $\pm$ 0.092 & 0.697 $\pm$ 0.035 & 0.992 $\pm$ 0.004 \\
& GD  & 0.431 $\pm$ 0.097 & 0.950 $\pm$ 0.014 & 0.556 $\pm$ 0.116 & 0.988 $\pm$ 0.008 \\
& UKL  & 0.835 $\pm$ 0.015 & 0.972 $\pm$ 0.017 & 0.928 $\pm$ 0.010 & 0.953 $\pm$ 0.061 \\
& DPO & 0.031 $\pm$ 0.019 & 0.841 $\pm$ 0.016 & 0.245 $\pm$ 0.007 & 0.924 $\pm$ 0.011 \\
& NPO & 0.487 $\pm$ 0.116 & 0.892 $\pm$ 0.081 & 0.720 $\pm$ 0.072 & 0.992 $\pm$ 0.001 \\
\midrule

\multirow{5}{*}{\textbf{Sparse}} 
& GA  & 0.536 $\pm$ 0.094 & 0.933 $\pm$ 0.016 & 0.713 $\pm$ 0.071 & 0.979 $\pm$ 0.004 \\
& GD  & 0.387 $\pm$ 0.028 & 0.894 $\pm$ 0.029 & 0.626 $\pm$ 0.000 & 0.971 $\pm$ 0.005 \\
& UKL  & 0.535 $\pm$ 0.465 & 0.718 $\pm$ 0.272 & 0.573 $\pm$ 0.427 & 0.769 $\pm$ 0.225 \\
& DPO & 0.035 $\pm$ 0.015 & 0.803 $\pm$ 0.038 & 0.226 $\pm$ 0.096 & 0.882 $\pm$ 0.027 \\
& NPO & 0.447 $\pm$ 0.007 & 0.913 $\pm$ 0.015 & 0.636 $\pm$ 0.060 & 0.978 $\pm$ 0.003 \\
\bottomrule
\end{tabular}
}
\end{table*}

\begin{table*}[ht]
\centering
\captionsetup{font=small,labelfont=bf}
\caption{Additional results of unlearning data of different domains in Dataset 1.}
\label{tab:additional_res_dataset1_domain}
\scalebox{0.7}{
\begin{tabular}{ll |cc |cc}
\toprule
{\textbf{Forget Data}} & 
{\textbf{Forget Method}} & 
{\textbf{Ind. Sales MRR}} & 
{\textbf{Ind. Emp. MRR}} & 
{\textbf{Ind. Sales THR}} & 
{\textbf{Ind. Emp. THR}} \\
\midrule
\multicolumn{6}{l}{\textbf{Llama2-7B}} \\

\multirow{5}{*}{\textbf{AC}} 
& GA  & 0.255 $\pm$ 0.002 & 0.331 $\pm$ 0.002 & 0.694 $\pm$ 0.010 & 0.726 $\pm$ 0.008 \\
& GD  & 0.252 $\pm$ 0.008 & 0.330 $\pm$ 0.001 & 0.702 $\pm$ 0.013 & 0.725 $\pm$ 0.003 \\
& KL  & 0.399 $\pm$ 0.043 & 0.417 $\pm$ 0.018 & 0.773 $\pm$ 0.053 & 0.802 $\pm$ 0.033 \\
& DPO & 0.280 $\pm$ 0.001 & 0.316 $\pm$ 0.000 & 0.626 $\pm$ 0.009 & 0.652 $\pm$ 0.008 \\
& NPO & 0.300 $\pm$ 0.005 & 0.314 $\pm$ 0.000 & 0.801 $\pm$ 0.004 & 0.815 $\pm$ 0.001 \\
\midrule
\multirow{5}{*}{\textbf{An}} 
& GA  & 0.279 $\pm$ 0.004 & 0.313 $\pm$ 0.006 & 0.673 $\pm$ 0.007 & 0.655 $\pm$ 0.007 \\
& GD  & 0.282 $\pm$ 0.002 & 0.322 $\pm$ 0.005 & 0.685 $\pm$ 0.005 & 0.683 $\pm$ 0.009 \\
& KL  & 0.366 $\pm$ 0.006 & 0.399 $\pm$ 0.002 & 0.806 $\pm$ 0.007 & 0.829 $\pm$ 0.012 \\
& DPO & 0.290 $\pm$ 0.001 & 0.329 $\pm$ 0.001 & 0.708 $\pm$ 0.004 & 0.712 $\pm$ 0.008 \\
& NPO & 0.304 $\pm$ 0.000 & 0.296 $\pm$ 0.000 & 0.787 $\pm$ 0.004 & 0.767 $\pm$ 0.006 \\
\midrule
\multicolumn{6}{l}{\textbf{Mistral-7B}} \\

\multirow{5}{*}{\textbf{AC}} 
& GA  & 0.277 $\pm$ 0.005 & 0.319 $\pm$ 0.005 & 0.670 $\pm$ 0.025 & 0.710 $\pm$ 0.022 \\
& GD  & 0.293 $\pm$ 0.003 & 0.324 $\pm$ 0.003 & 0.699 $\pm$ 0.006 & 0.731 $\pm$ 0.015 \\
& KL  & 0.342 $\pm$ 0.003 & 0.343 $\pm$ 0.002 & 0.805 $\pm$ 0.002 & 0.798 $\pm$ 0.002 \\
& DPO & 0.303 $\pm$ 0.003 & 0.329 $\pm$ 0.001 & 0.732 $\pm$ 0.007 & 0.757 $\pm$ 0.001 \\
& NPO & 0.352 $\pm$ 0.000 & 0.345 $\pm$ 0.001 & 0.811 $\pm$ 0.003 & 0.748 $\pm$ 0.001 \\
\midrule
\multirow{5}{*}{\textbf{An}} 
& GA  & 0.322 $\pm$ 0.005 & 0.325 $\pm$ 0.002 & 0.757 $\pm$ 0.006 & 0.698 $\pm$ 0.010 \\
& GD  & 0.321 $\pm$ 0.002 & 0.324 $\pm$ 0.003 & 0.808 $\pm$ 0.012 & 0.745 $\pm$ 0.014 \\
& KL  & 0.351 $\pm$ 0.002 & 0.346 $\pm$ 0.008 & 0.815 $\pm$ 0.002 & 0.808 $\pm$ 0.003 \\
& DPO & 0.310 $\pm$ 0.001 & 0.320 $\pm$ 0.001 & 0.730 $\pm$ 0.005 & 0.740 $\pm$ 0.001 \\
& NPO & 0.301 $\pm$ 0.001 & 0.293 $\pm$ 0.003 & 0.615 $\pm$ 0.037 & 0.537 $\pm$ 0.030 \\
\midrule
\multicolumn{6}{l}{\textbf{{Gemma-7B}}} \\

\multirow{5}{*}{\textbf{AC}} 
& GA  & 0.980 $\pm$ 0.003 & 0.997 $\pm$ 0.002 & 1.000 $\pm$ 0.000 & 1.000 $\pm$ 0.000 \\
& GD  & 0.957 $\pm$ 0.009 & 0.992 $\pm$ 0.004 & 0.994 $\pm$ 0.003 & 0.996 $\pm$ 0.003 \\
& KL  & 0.958 $\pm$ 0.006 & 0.994 $\pm$ 0.006 & 0.969 $\pm$ 0.005 & 0.995 $\pm$ 0.006 \\
& DPO & 0.717 $\pm$ 0.056 & 0.953 $\pm$ 0.043 & 0.841 $\pm$ 0.042 & 0.980 $\pm$ 0.020 \\
& NPO & 0.983 $\pm$ 0.002 & 1.000 $\pm$ 0.001 & 0.999 $\pm$ 0.002 & 1.000 $\pm$ 0.001 \\
\midrule
\multirow{5}{*}{\textbf{An}} 
& GA  & 0.999 $\pm$ 0.001 & 0.849 $\pm$ 0.000 & 0.999 $\pm$ 0.001 & 0.920 $\pm$ 0.000 \\
& GD  & 0.987 $\pm$ 0.004 & 0.903 $\pm$ 0.018 & 0.995 $\pm$ 0.003 & 0.964 $\pm$ 0.010 \\
& KL  & 0.954 $\pm$ 0.030 & 0.848 $\pm$ 0.013 & 0.973 $\pm$ 0.016 & 0.935 $\pm$ 0.012 \\
& DPO & 0.929 $\pm$ 0.010 & 0.831 $\pm$ 0.015 & 0.957 $\pm$ 0.005 & 0.918 $\pm$ 0.008 \\
& NPO & 0.990 $\pm$ 0.007 & 0.854 $\pm$ 0.000 & 0.995 $\pm$ 0.003 & 0.930 $\pm$ 0.006 \\
\bottomrule
\end{tabular}
}
\end{table*}

\begin{table*}[ht]
\centering
\captionsetup{font=small,labelfont=bf}
\caption{Additional results of unlearning pre-training data with different levels of inter-connectivity.}
\label{tab:unlearning_entities}
\scalebox{0.7}{
\begin{tabular}{ll |cc |cc}
\toprule
{\textbf{Forget Entity}} & 
{\textbf{Forget Method}} & 
{\textbf{Forget Set MRR}} & 
{\textbf{Retain Set MRR}} & 
{\textbf{Forget Set THR}} & 
{\textbf{Retain Set THR}} \\
\midrule
\multicolumn{6}{l}{\textbf{Llama2-7B}} \\

\multirow{5}{*}{\shortstack[l]{\textbf{Goldman} \\ \textbf{Sachs}}} 
& GA  & 0.185 $\pm$ 0.017 & 0.277 $\pm$ 0.001 & 0.513 $\pm$ 0.031 & 0.566 $\pm$ 0.008 \\
& GD  & 0.188 $\pm$ 0.025 & 0.277 $\pm$ 0.001 & 0.512 $\pm$ 0.042 & 0.564 $\pm$ 0.005 \\
& UKL  & 0.266 $\pm$ 0.089 & 0.365 $\pm$ 0.006 & 0.432 $\pm$ 0.078 & 0.596 $\pm$ 0.032 \\
& DPO & 0.191 $\pm$ 0.068 & 0.278 $\pm$ 0.003 & 0.459 $\pm$ 0.123 & 0.561 $\pm$ 0.010 \\
& NPO & 0.113 $\pm$ 0.010 &	0.277 $\pm$ 0.001 &	0.443 $\pm$ 0.030 &	0.572 $\pm$ 0.007\\

\midrule
\multirow{5}{*}{\shortstack[l]{\textbf{William} \\ \textbf{Shakespeare}}} 
& GA  & 0.228 $\pm$ 0.045 & 0.273 $\pm$ 0.006 & 0.504 $\pm$ 0.070 & 0.557 $\pm$ 0.011 \\
& GD  & 0.240 $\pm$ 0.023 & 0.277 $\pm$ 0.002 & 0.524 $\pm$ 0.077 & 0.565 $\pm$ 0.006 \\
& UKL  & 0.311 $\pm$ 0.058 & 0.362 $\pm$ 0.015 & 0.567 $\pm$ 0.081 & 0.630 $\pm$ 0.048 \\
& DPO & 0.240 $\pm$ 0.034 & 0.276 $\pm$ 0.002 & 0.472 $\pm$ 0.079 & 0.555 $\pm$ 0.007 \\
& NPO & 0.157 $\pm$ 0.051 &	0.277 $\pm$ 0.001	& 0.391 $\pm$ 0.019 &	0.548 $\pm$ 0.007 \\
\midrule
\multicolumn{6}{l}{\textbf{Mistral-7B}} \\

\multirow{5}{*}{\shortstack[l]{\textbf{Goldman} \\ \textbf{Sachs}}} 
& GA  & 0.273 $\pm$ 0.011 & 0.344 $\pm$ 0.004 & 0.598 $\pm$ 0.013 & 0.732 $\pm$ 0.016 \\
& GD  & 0.262 $\pm$ 0.011 & 0.345 $\pm$ 0.004 & 0.605 $\pm$ 0.032 & 0.733 $\pm$ 0.018 \\
& UKL  & 0.356 $\pm$ 0.039 & 0.337 $\pm$ 0.003 & 0.735 $\pm$ 0.021 & 0.776 $\pm$ 0.012 \\
& DPO & 0.045 $\pm$ 0.026 & 0.331 $\pm$ 0.001 & 0.120 $\pm$ 0.059 & 0.703 $\pm$ 0.009 \\
& NPO & 0.233 $\pm$ 0.041 & 0.337 $\pm$ 0.000 & 0.513 $\pm$ 0.017 & 0.695 $\pm$ 0.006 \\
\midrule
\multirow{5}{*}{\shortstack[l]{\textbf{William} \\ \textbf{Shakespeare}}} 
& GA  & 0.237 $\pm$ 0.049 & 0.323 $\pm$ 0.007 & 0.469 $\pm$ 0.080 & 0.689 $\pm$ 0.023 \\
& GD  & 0.255 $\pm$ 0.016 & 0.329 $\pm$ 0.003 & 0.557 $\pm$ 0.065 & 0.706 $\pm$ 0.016 \\
& UKL  & 0.337 $\pm$ 0.070 & 0.321 $\pm$ 0.005 & 0.714 $\pm$ 0.030 & 0.788 $\pm$ 0.005 \\
& DPO & 0.035 $\pm$ 0.007 & 0.318 $\pm$ 0.001 & 0.133 $\pm$ 0.019 & 0.716 $\pm$ 0.008 \\
& NPO & 0.259 $\pm$ 0.009 & 0.331 $\pm$ 0.000 & 0.606 $\pm$ 0.018 & 0.718 $\pm$ 0.001 \\
\midrule
\multicolumn{6}{l}{\textbf{Gemma-7B}} \\

\multirow{5}{*}{\shortstack[l]{\textbf{Goldman} \\ \textbf{Sachs}}} 
& GA & 0.737 $\pm$ 0.051 & 0.993 $\pm$ 0.006 & 0.799 $\pm$ 0.026 & 1.000 $\pm$ 0.000 \\
& GD & 0.652 $\pm$ 0.073 & 0.990 $\pm$ 0.009 & 0.727 $\pm$ 0.077 & 0.996 $\pm$ 0.003\\
& UKL  & 0.814 $\pm$ 0.162 & 0.981 $\pm$ 0.022 & 0.910 $\pm$ 0.086 & 0.991 $\pm$ 0.011 \\
& DPO & 0.532 $\pm$ 0.032 & 1.000 $\pm$ 0.000 & 0.756 $\pm$ 0.052 & 1.000 $\pm$ 0.000 \\
& NPO & 0.579  $\pm$ 0.098	& 0.988 $\pm$ 0.008 & 0.688  $\pm$ 0.085 & 0.994 ± 0.003 \\
\midrule
\multirow{5}{*}{\shortstack[l]{\textbf{William} \\ \textbf{Shakespeare}}} 
& GA & 0.633 $\pm$ 0.041 & 0.984 $\pm$ 0.003 & 0.773 $\pm$ 0.045 & 0.996 $\pm$ 0.003 \\
& GD & 0.559 $\pm$ 0.131 & 0.962 $\pm$ 0.034 & 0.705 $\pm$ 0.096 & 0.986 $\pm$ 0.015\\
& UKL  & 0.642 $\pm$ 0.246 & 0.863 $\pm$ 0.134 & 0.742 $\pm$ 0.223 & 0.929 $\pm$ 0.071 \\
& DPO & 0.051 $\pm$ 0.039 & 0.887 $\pm$ 0.006 & 0.174 $\pm$ 0.066 & 0.940 $\pm$ 0.007 \\
& NPO & 0.587 $\pm$ 0.117 & 0.976 $\pm$ 0.011 &	0.722 $\pm$ 0.079 &	0.989 $\pm$ 0.009\\
\bottomrule
\end{tabular}
}
\end{table*}

\end{document}